\documentclass[lettersize,journal]{IEEEtran}
\pdfoutput=1
\usepackage{amsmath,amsfonts}
\usepackage{algorithmic}
\usepackage{algorithm}
\usepackage{array}
\usepackage[caption=false,font=footnotesize,labelfont=rm,textfont=rm]{subfig}
\usepackage{textcomp}
\usepackage{stfloats}
\usepackage{url}
\usepackage{verbatim}
\usepackage{graphicx}
\usepackage{cite}
\usepackage{bbding}
\usepackage{booktabs}
\usepackage{multirow}
\usepackage{color}

\begin{document}

\title{DEA-Net: Single image dehazing based on detail-enhanced convolution and content-guided attention}

\author{Zixuan Chen$^\dagger$, Zewei He$^\dagger$, Zhe-Ming Lu$^\ast$,~\IEEEmembership{senior member,~IEEE}
	\thanks{Z. Chen, Z. He and Z.M. Lu are with School of Aeronautics and Astronautics, Zhejiang University (e-mail: zeweihe@zju.edu.cn).}
	\thanks{This work was funded by China Postdoctoral Science Foundation funded project (No. 2022M712792).}
	\thanks{$^\dagger$The first two authors contribute equally to this work.}
	\thanks{$^\ast$Corresponding author: Zhe-Ming Lu.}
	\thanks{This paper was produced by the IEEE Publication Technology Group. They are in Piscataway, NJ.}
	\thanks{Manuscript received April 19, 2021; revised August 16, 2021.}}

\markboth{Journal of \LaTeX\ Class Files,~Vol.~14, No.~8, August~2021}%
{Shell \MakeLowercase{\textit{et al.}}: A Sample Article Using IEEEtran.cls for IEEE Journals}



\maketitle

\newcommand{\etal}{et al.}
\newcommand{\eg}{e.g., }
\newcommand{\ie}{i.e., }

\begin{abstract}
Single image dehazing is a challenging ill-posed problem which estimates latent haze-free images from observed hazy images.
Some existing deep learning based methods are devoted to improving the model performance via increasing the depth or width of convolution.
The learning ability of convolutional neural network (CNN) structure is still under-explored.
In this paper, a detail-enhanced attention block (DEAB) consisting of the detail-enhanced convolution (DEConv) and the content-guided attention (CGA) is proposed to boost the feature learning for improving the dehazing performance.
Specifically, the DEConv integrates prior information into normal convolution layer to enhance the representation and generalization capacity. 
Then by using the re-parameterization technique, DEConv is equivalently converted into a vanilla convolution with NO extra parameters and computational cost.
By assigning unique spatial importance map (SIM) to every channel, CGA can attend more useful information encoded in features.
In addition, a CGA-based mixup fusion scheme is presented to effectively fuse the features and aid the gradient flow.
By combining above mentioned components, we propose our detail-enhanced attention network (DEA-Net) for recovering high-quality haze-free images.
Extensive experimental results demonstrate the effectiveness of our DEA-Net, outperforming the state-of-the-art (SOTA) methods by boosting the PSNR index over 41 dB with only 3.653 M parameters. The source code of our DEA-Net will be made available at https://github.com/cecret3350/DEA-Net.

\end{abstract}

\begin{IEEEkeywords}
Image dehazing, Detail-enhanced convolution, Content-guided attention, Fusion scheme.
\end{IEEEkeywords}

\section{Introduction}
\label{sec: introduction}
\IEEEPARstart{I}{mages}
captured under hazy scenes usually suffer from noticeable visual quality degradation in contrast or color distortion \cite{Tan2008CVPR}, leading to significant performance drop when inputting to some high-level vision tasks (e.g., object detection, semantic segmentation).
Haze-free images are highly demanded or required among these tasks.
Therefore, single image dehazing, which aims to recover the clean scene from the corresponding hazy image, has attracted significant attention among both the academic and industrial communities over the past decade.
As a fundamental low-level image restoration task, image dehazing can be the pre-processing step of the subsequent high-level vision tasks.
In this paper, we attempt to develop an effective algorithm to remove the haze and recover the details from the hazy input.

Recently, with the rapid development of deep learning, convolution neural network (CNN) based dehazing methods achieve superior performance \cite{cai2016TIP,li2017ICCV,li2020TIP,qin2020AAAI,wu2021CVPR}.
Earlier CNN-based methods \cite{cai2016TIP,ren2016ECCV,zhang2018CVPR} first estimate the transmission map and the atmospheric light separately, and then utilize the atmospheric scattering model (ASM) \cite{Narasimhan2003TPAMI-ASM} to derive the haze-free images.
Typically, the transmission map is supervised by the ground truth, which is used for synthesizing the training dataset.
However, inaccurate estimation of the transmission map or the atmospheric light would significantly influence the image restoration results.
More recently, some methods \cite{dong2020CVPR,wu2021CVPR,ye2022ECCVORAL} prefer to predict the latent haze-free images in an end-to-end manner since it tends to achieve promising results.

\begin{figure}[!t]
	\centering
	\includegraphics[width=0.48\textwidth]{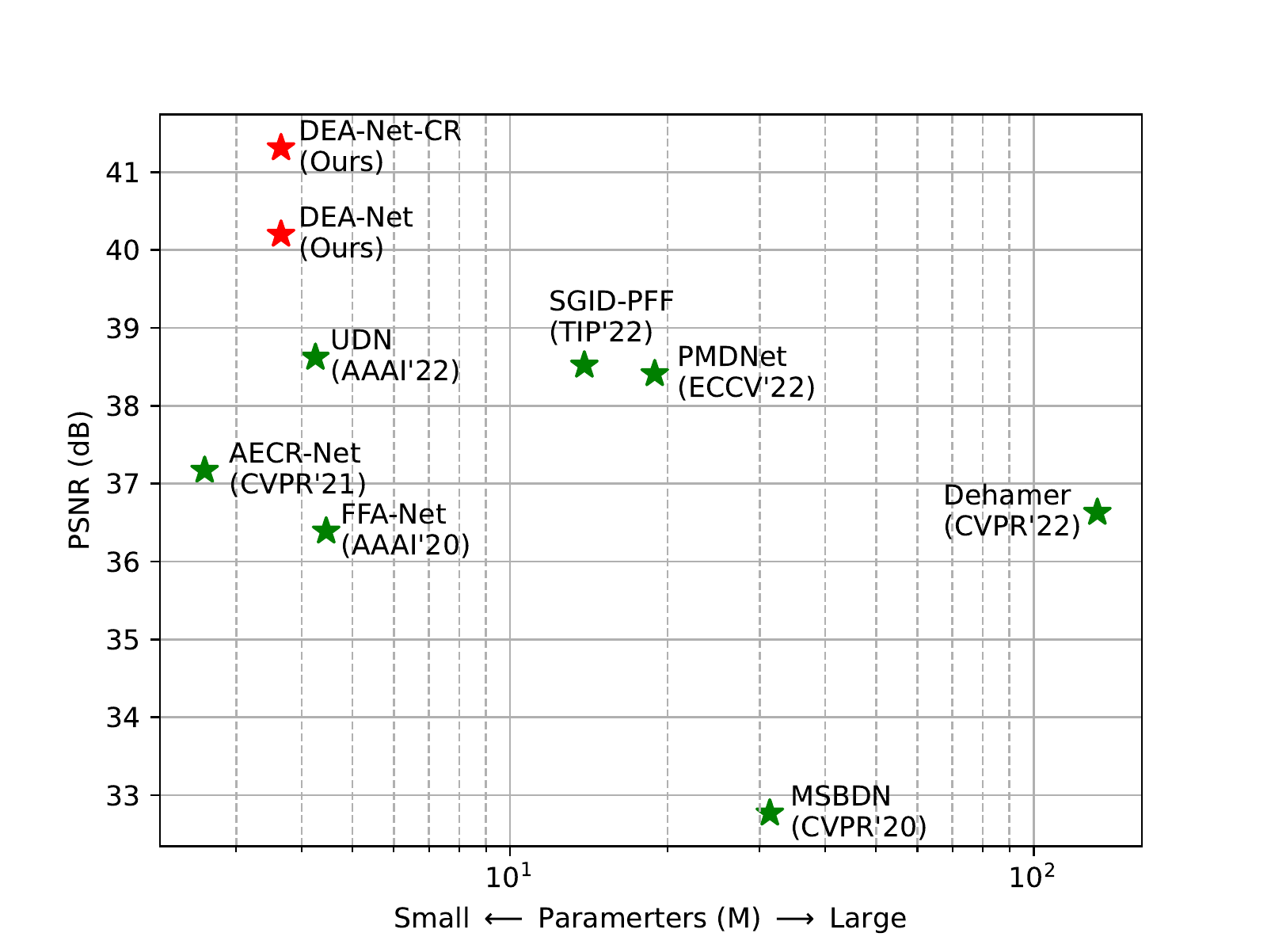}
	\caption{Graph of PSNR vs. number of parameters. We compare our DEA-Net with some state-of-the-art methods (after 2020). The results are tested on SOTS-indoor dataset. Note that AECR-Net adopts a sharing strategy to reduce the number of parameters.}
	\label{fig:fig1}
\end{figure}

However, there still exists two main issues:

(1) \emph{Less effectiveness of vanilla convolution.}
Previous works \cite{he2010TPAMI,berman2016CVPR,zhu2015TIP} prove that well-designed priors like dark channel prior \cite{He2009CVPR-DCP,he2010TPAMI}, non-local haze-line prior \cite{berman2016CVPR}, and color attenuation prior \cite{zhu2015TIP}, are helpful for recovering missing information.
Most of existing dehazing methods \cite{qin2020AAAI,Wu2020CVPRW,wu2021CVPR} adopt classical convolution layers for feature extraction without utilizing these priors. 
However, vanilla convolutions search the vast solution space without any constrains, which to some extend may limit the expressive ability (or modeling capacity).
In addition, some transformer-based methods \cite{guo2022CVPR} expand the receptive field to the whole image for mining long-distance dependencies.
They can enhance the expressive ability (or modeling capacity) at the cost of complex training strategy and tedious hyper-parameter tuning.
Also, the prohibitive computational cost and vast GPU memory occupation cannot be ignored. 
In this regard, the ideal solution is to embed the well-designed priors into CNN for improving the feature learning capability.

(2) \emph{Haze non-uniformity.}
There are two kinds of non-uniformity in the dehazing problem: uneven haze distribution in image level and channel-wise haze difference in feature level.
To cope with the first one, Qin \etal \cite{qin2020AAAI} employed a pixel attention (i.e., spatial attention) to generate a spatial importance map (SIM), which can adaptively indicating the importance levels of different pixel locations. 
Through this discriminative strategy, the FFA-Net model treats thin and thick haze regions unequally.
Similarly, Ye \etal \cite{ye2022ECCVORAL} tried to model the density of haze distribution via a density estimation module, which essentially is also a spatial attention.
However, seldom researchers paid attention to the non-uniformity in feature level, which remains to be unexploited.
The channel attention used in \cite{qin2020AAAI} can produce a channel-wise attention vector \footnote{The global average pooling (GAP) operation reduces the spatial dimensions to one point.} to indicate the importance level of each channel, which fails to consider the contextual information in the spatial dimensions.
The haze information is encoded into the feature maps after applying convolution layers.
Different channels in the feature space have different meanings depending on the role of the filters applied.
In this regard, we argue that spatial importance maps should be channel-specific, and consider two kinds of non-uniformity (image level and feature level) simultaneously.

To address above mentioned issues, we design a detail-enhanced attention block (DEAB), which consists of a detail-enhanced convolution (DEConv) and a content-guided attention (CGA) mechanism.
The DEConv contains five convolution layers (four difference convolutions \cite{Yu2020CVPR-CDC} and one vanilla convolution), which are parallel deployed for feature extraction.
Specifically, a central difference convolution (CDC), an angular difference convolution (ADC), a horizontal difference convolution (HDC), and a vertical difference convolution (VDC) are adopted to integrate traditional local descriptors into the convolution layer, thus can enhance the representation and generalization capacity.
In difference convolutions, the pixel differences in the image are firstly calculated, and then convolved with the convolution kernel to generate the output feature maps.
The strategy of pixel pair's difference calculation can be designed to explicitly encode prior information into CNN.
For instance, HDC and VDC explicitly encode the gradient prior into the convolution layers via learning beneficial gradient information.

Moreover, the sophisticated attention mechanism (i.e., CGA) is a two-step attention generator, which can produce the coarse spatial attention map firstly and then refine it to the fine version.
Specifically, given certain input feature maps, we utilize the spatial attention mechanism presented in \cite{woo2018ECCV} and the channel attention presented in \cite{Hu2018CVPR-SE} to generate the initial SIMs (i.e., the coarse version).
Then, the initial SIMs are refined according to every channel of input feature maps to produce final SIMs.
By using the content of input features to guide the generation of SIMs, CGA can focus on the unique part of features in each channel.
It is worth mentioning that CGA as a universal basic block can be plug into neural networks to improve the performance in various image restoration tasks.

Besides the improvements mentioned above, we re-parameterize the learned kernel weights of the parallel convolutions to reduce the number of parameters and accelerate the training the testing process.
The five parallel convolutions are simplified into one vanilla convolution layer with applying some constrains to the kernel weights and by using the linear property of convolution layers.
Therefore, the proposed DEConv can extract richful features for improving dehazing performance while keeping the number of parameters and computational cost equal to the vanilla convolution.
Fig.~\ref{fig:fig1} shows the efficiency and effectiveness of our method.


Following \cite{dong2020CVPR,bai2022TIP,wu2021CVPR,hong2022AAAI}, we also adopt a U-net-like framework to make the major time-consuming convolution computations in the low-resolution space. Among them, the fusion of shallow and deep features is widely used.
Feature fusion can enhance the information flow from shallow layers to deep ones, which is effective for feature preserving and gradient back-propagation.
The information encoded in the shallow features is tremendously different from the information encoded in the deep features, since the diverse receptive fields. One single pixel in the deep features are originated from a region of pixels in the shallow features.
Simple addition or concatenation operation is unable to solve the receptive field mismatch problem.
We further propose a CGA-based mixup scheme to adaptively fuse the low-level features in the encoder part with corresponding high-level features, by modulating the features via learned spatial weights.

\begin{figure*}[!t]
	\centering
	\includegraphics[width=0.95\linewidth]{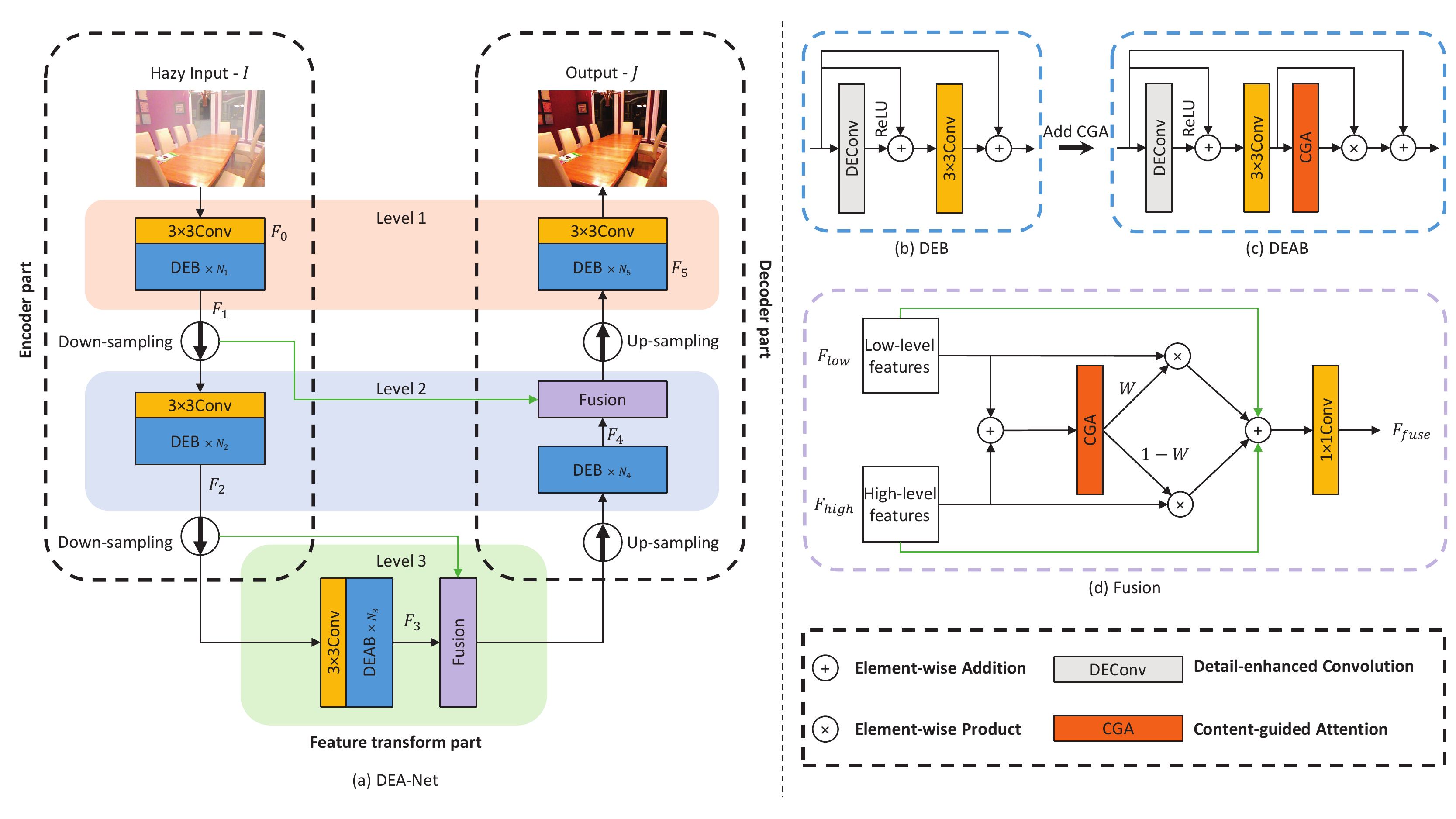}
	\caption{The overall architecture of our proposed detail-enhanced attention network (DEA-Net), which is a three-level encoder-decoder-like architecture. DEA-Net contains three parts: encoder part, feature transform part, and decoder part. We deploy detail-enhanced attention blocks (DEABs) in the feature transform part, and deploy detail-enhanced blocks (DEBs) in rest parts.}
	\label{fig:overall structure}
\end{figure*}

The diagram of our proposed method are shown in Fig.~\ref{fig:overall structure}. 
We term the proposed single image dehazing model as DEA-Net by introducing the \textbf{D}etail-\textbf{E}nhanced \textbf{A}ttention block (DEAB) with the \textbf{D}etail-\textbf{E}nhanced convolution and the content-guided \textbf{A}ttention.

To conclude, we have following main contributions:

\begin{itemize}
	\item We design a detail-enhanced convolution (DEConv), which contains parallel vanilla and difference convolutions. To the best of our knowledge, it's the first time that difference convolutions are introduced to solve the image dehazing problem. By encoding prior information into normal convolution layer, the representation and generalization capacity of DEConv is enhanced for improving dehazing performance. In addition, we equivalently convert the DEConv into a normal convolution with \textbf{NO} extra parameters and computational cost via using re-parameterization technique.
	\item We propose a novel attention mechanism called content-guided attention (CGA) to generate the channel-specific SIMs in a coarse-to-fine manner. By using input features to guide the generation of SIMs, CGA assigns unique SIM to every channel, making the model attend significant regions of each channel. Thus, more useful information encoded in features can be emphasized to effectively improve the performance. Moreover, a CGA-based mixup fusion scheme is presented to effectively fuse the low-level features in the encoder part with corresponding high-level features.
	\item By combining DEConv and CGA, and using CGA-based mixup fusion scheme, we propose our detail-enhanced attention network (DEA-Net) for reconstructing high-quality haze-free images. DEA-Net shows superior performance over the state-of-the-art dehazing methods on multiple benchmark datasets, achieving more accurate results with faster inference speed.
\end{itemize}

The remainder of this paper is organized as follows. We first review a number of deep learning-based dehazing methods in Sec.~\ref{sec: related work}.
Sec.~\ref{sec: methodology} describes the proposed EDA-Net model in detail, and Sec.~\ref{sec: experiment} shows the experimental results.
Finally, Sec.~\ref{sec: conclusion} concludes this paper.

{\color{black}
\section{Related Work}
\label{sec: related work}
\subsection{Single Image Dehazing}

For single image dehazing, existing methods can be mainly divided into two categories. 
One is to manually generalize the statistical discrepancy between the hazy and haze-free images as empirical priors. 
Another one aims to directly or indirectly learning the mapping function based on large-scale datasets. 
We usually term the former as the prior-based methods and the latter as the data-driven methods.

The prior-based methods are the pioneers of image dehazing. 
They usually rely on atmospheric scattering model (ASM) \cite{Narasimhan2003TPAMI-ASM} and handcraft priors. 
The widely known priors include dark channel prior (DCP) \cite{He2009CVPR-DCP,he2010TPAMI}, non-local prior (NLP) \cite{berman2016CVPR}, color attenuation prior (CAP) \cite{zhu2015TIP}, etc. 
He et al. \cite{He2009CVPR-DCP,he2010TPAMI} proposed DCP based on a key observation - most local patches in haze-free outdoor images contain some pixels which have very low intensities in at least one color channel, which can help estimate the transmission map.
CAP \cite{zhu2015TIP} starts from the HSV color model, and establishes a linear relationship between depth and the difference of brightness and saturation.
Berman et al. \cite{berman2016CVPR} found that pixel clusters of haze-free images will become haze-lines when haze presents. 
These prior-based methods have achieved promising dehazing results.
However, they tend to work well only in specific scenes which happen to satisfy their assumptions.

Recently, with the rising of deep learning, researchers focused on data-driven methods, since they can achieve better performance.
Earlier data-driven methods usually perform dehazing based on the physical model. 
For instance, DehazeNet \cite{cai2016TIP} and MSCNN \cite{ren2016ECCV} utilize CNNs to estimate the transmission map. 
Then, AOD-Net \cite{li2017ICCV} rewrites the ASM and estimates atmospheric light together with transmission map. 
Later, DCPDN \cite{zhang2018CVPR} estimates the transmission map and atmospheric light by two different networks. 
However, the cumulative errors introduced by inaccurate estimations of transmission map and atmospheric light may cause the performance degradation. 

To avoid this, more recent works tend to recover the haze-free image from the hazy image directly without the help of the physical model. 
GFN \cite{ren2018CVPR} gates and fuses three enhanced images from original hazy inputs to generate the haze-free images. 
GridDehazeNet \cite{liu2019ICCV} utilizes a three-stage attention-based grid network to recover the haze-free images. 
MSBDN \cite{dong2020CVPR} utilizes boosting strategy and back-projection technique to enhance the feature fusion. 
FFA-Net \cite{qin2020AAAI} introduces the feature attention mechanism (FAM) to dehazing network to deal with different types of information. 
AECR-Net \cite{wu2021CVPR} reuses the feature attention block (FAB) \cite{qin2020AAAI} and proposes a novel contrastive regularization, which can benefit from both positive samples and negative samples. 
UDN \cite {hong2022AAAI} analyzes two types of uncertainty in image dehazing, and utilizes them to increase the dehazing performance. 
PMDNet \cite{ye2022ECCVORAL} and Dehamer \cite{guo2022CVPR} adopt transformer to build long-range dependencies and perform dehazing with the guidance of haze density. 
However, as data-driven methods develop and dehazing performance improves, the complexity of dehazing networks also increases. 
Different from previous works, we rethink the deficiencies of vanilla convolution in image dehazing and design a novel convolution operator by combining well-designed priors into CNN for improving the feature learning capability. 
We also dig deeper into the unexploited non-uniformity of haze in feature level. 

\subsection{Difference Convolution}
The origin of difference convolutions can be traced back to the local binary pattern (LBP) \cite{ojala2002TPAMI}, which encodes the pixel differences in the local patch to a decimal number for texture classification. 
Since the success of CNNs in computer vision tasks, Xu et al. \cite{juefei2017CVPR} proposed the local binary convolution (LBC) which encodes the pixel differences by using non-linear activation functions and linear convolution layers. 
Recently, Yu et al. \cite{yu2020CVPR} proposed the central difference convolution (CDC) to directly encode the pixel differences with completely learnable weights. 
Later, various forms of difference convolutions have been proposed, such as cross central difference convolution \cite{yu2021IJCAI} and pixel difference convolution \cite{su2021ICCV}. 
Considering the nature of the difference convolution for capturing gradient-level information, we firstly introduce it to single image dehazing for improving the performance. 
}


\section{Methodology}
\label{sec: methodology}
As shown in Fig.~\ref{fig:overall structure}, our DEA-Net consists of three parts: encoder part, feature transform part, and decoder part.
As the core of our DEA-Net, the feature transform part adopts stacked detail-enhanced attention blocks (DEABs) to learn haze-free features.
There are three levels in the hierarchical structure, and we employ different blocks in different levels to extract corresponding features (level 1\&2: DEB, level 3: DEAB).
Given a hazy input image $I\in \mathbb{R}^{3\times H\times W}$, the goal of DEA-Net is to restore the corresponding haze-free image $J\in \mathbb{R}^{3\times H\times W}$. 

\subsection{Detail-enhanced Convolution}
\begin{figure}[!t]
	\centering
	\includegraphics[width=0.7\linewidth]{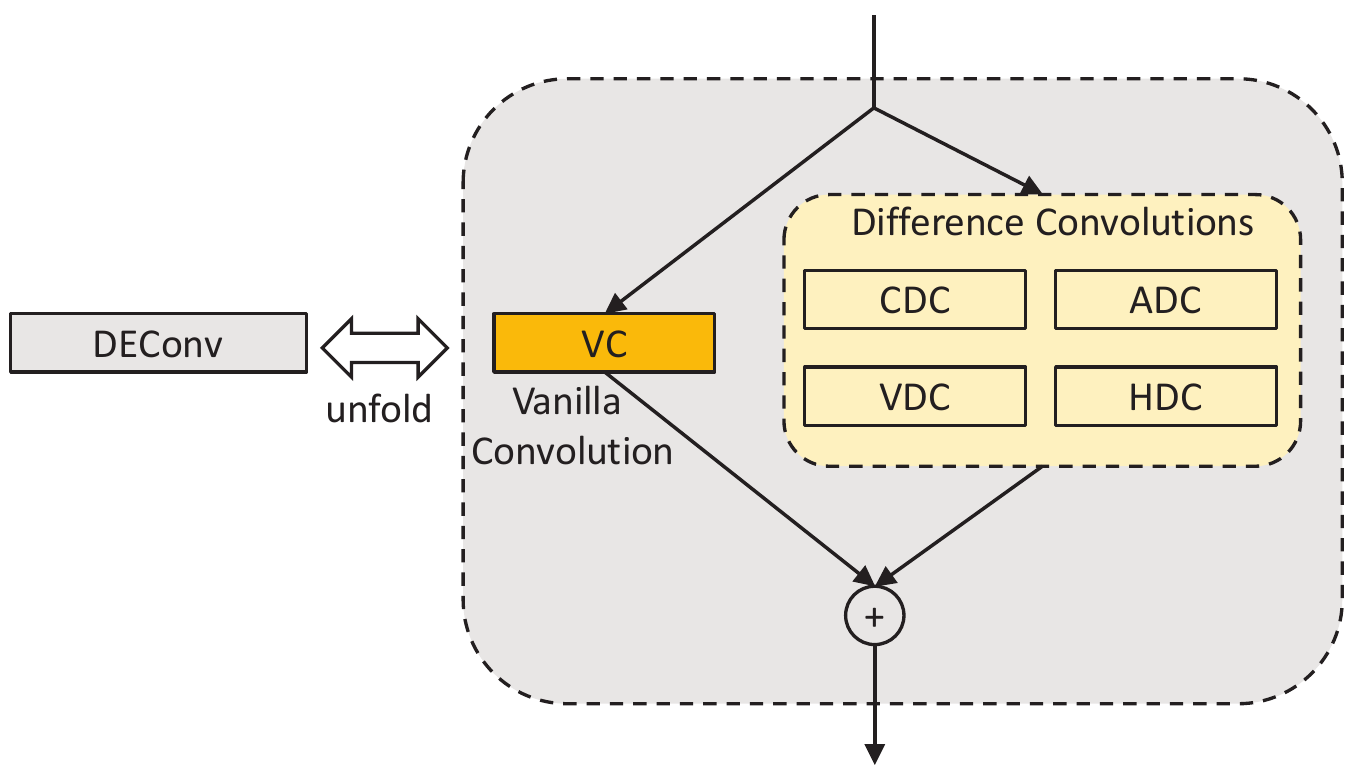}
	\caption{Detail-enhanced convolution (DEConv). It contains five parallel deployed convolution layers including: a vanilla convolution (VC), a central difference convolution (CDC), an angular difference convolution (ADC), a horizontal difference convolution (HDC), and a vertical difference convolution (VDC).}
	\label{fig:fig3}
\end{figure}

In single image dehazing domain, previous methods \cite{qin2020AAAI,Wu2020CVPRW,wu2021CVPR} usually utilize vanilla convolution (VC) layers for feature extraction and learning.
Normal convolution layers search the vast solution space without any constrains (even start from random initialization), restricting the expressive ability or modeling capacity.
Then we notice that the high-frequency information (\eg edges and contours) is of great significance in recovering an image captured under the hazy scene.
Based on this, some researchers \cite{zhang2018CVPR,bai2022TIP,wang2021KBS} adopted the edge prior in the dehazing model to help restore sharper contours. 
Inspired by their works \cite{zhang2018CVPR,wang2021KBS}, we design a detail-enhanced convolution (DEConv) layer (see in Fig.~\ref{fig:fig3}), which can integrate well-designed priors into vanilla convolution layers.

Before elaborating the proposed DEConv in detail, we first recap the difference convolution (DC).
Previous works \cite{yu2020CVPR,yu2020TPAMI,su2021ICCV,yu2021IJCAI} usually describe the difference convolution as the convolution of pixel differences (the pixel differences are firstly calculated, and then convolved with the kernel weights to generate feature maps), which can enhance the representation and generalization capacity of vanilla convolution.
Central difference convolution (CDC) and angular difference convolution (ADC) are two kinds of typical DCs, and implemented by re-arranging learned kernel weights to save computational cost and memory consumption \cite{su2021ICCV}.
It proves to be effective for edge detection \cite{su2021ICCV} and face anti-spoofing tasks \cite{yu2020CVPR,yu2020TPAMI,yu2021IJCAI}.
\textbf{To the best of our knowledge, it is the first time that we introduce DC to solve the single image dehazing problem. }

\begin{figure}[!t]
	\centering
	\includegraphics[width=0.8\linewidth]{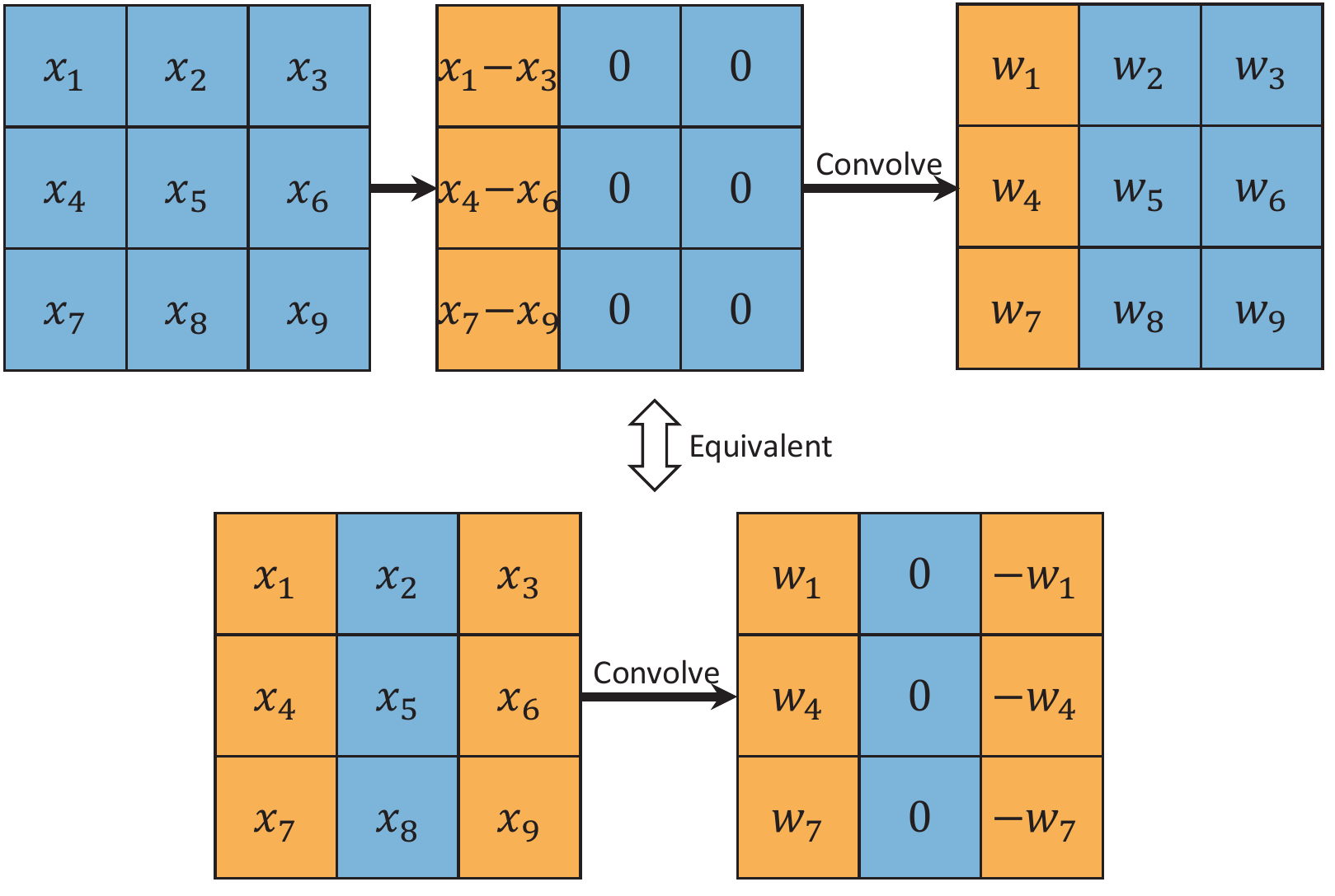}
	\caption{The derivation of horizontal difference convolution (HDC).}
	\label{fig:HDC}
\end{figure}

In our implementation, we employ five convolution layers (four DCs \cite{Yu2020CVPR-CDC} and one vanilla convolution), which are parallel deployed for feature extraction.
In DCs, the strategy of pixel pair's difference calculation can be designed to explicitly encode prior information into CNN.
For our DEConv, besides the central difference convolution (CDC) and the angular difference convolution (ADC), we derive the horizontal difference convolution (HDC) and the vertical difference convolution (VDC) to integrate traditional local descriptors (like Sobel \cite{sobel19683x3}, Prewitt \cite{prewitt1970object}, or Scharr \cite{scharr2000optimal}) into the convolution layer.
As shown in Fig.~\ref{fig:HDC}, taking HDC as an example, the horizontal gradient is firstly calculated by computing the differences of selected pixel pairs. 
After training, we re-arrange the learned kernel weights equivalently, and apply convolution directly to the untouched input features.
Note that, the equivalent kernel has the similar format of traditional local descriptors (the sum of horizontal weights equals to zero).
Horizontal kernels of Sobel \cite{sobel19683x3}, Prewitt \cite{prewitt1970object}, and Scharr \cite{scharr2000optimal} can be regarded as the special case of the equivalent kernel.
VDC has the similar derivation by changing the horizontal gradient to corresponding vertical counterpart.
Both HDC and VDC explicitly encode the gradient prior into the convolution layers to enhance the representation and generalization capacity via learning beneficial gradient information.

In our design, the vanilla convolution serves to obtain the intensity-level information while the difference convolutions are used to enhance gradient-level information.
We simply add the learned features together to obtain the output of DEConv.
We trust more sophisticated designs of the way for calculating the pixel difference can further benefit image restoration task, which is not the main direction of this paper.

However, deploying five parallel convolution layers for feature extraction will undesirably cause the increase of parameters and inference time.
We seek to exploit the additivity of convolution layers for simplifying the parallel deployed convolutions into a single standard convolution.
We notice a useful property of the convolution: 
if several 2D kernels with the identical size operate on the same input with the same stride and padding to produce outputs, and their outputs are summed up to obtain the final output, we can add up these kernels on the corresponding positions to obtain an equivalent kernel which will produce the identical final output.
Surprisingly, our DEConv exactly fits this situation.
Given the input features $F_{in}$, DEConv can output $F_{out}$ with identical computational cost and inference time to a vanilla convolution layer by utilizing re-parameterization technique.
The formula is as follows (the biases are omitted for simplification):

\begin{figure}[!t]
	\centering
	\includegraphics[width=0.9\linewidth]{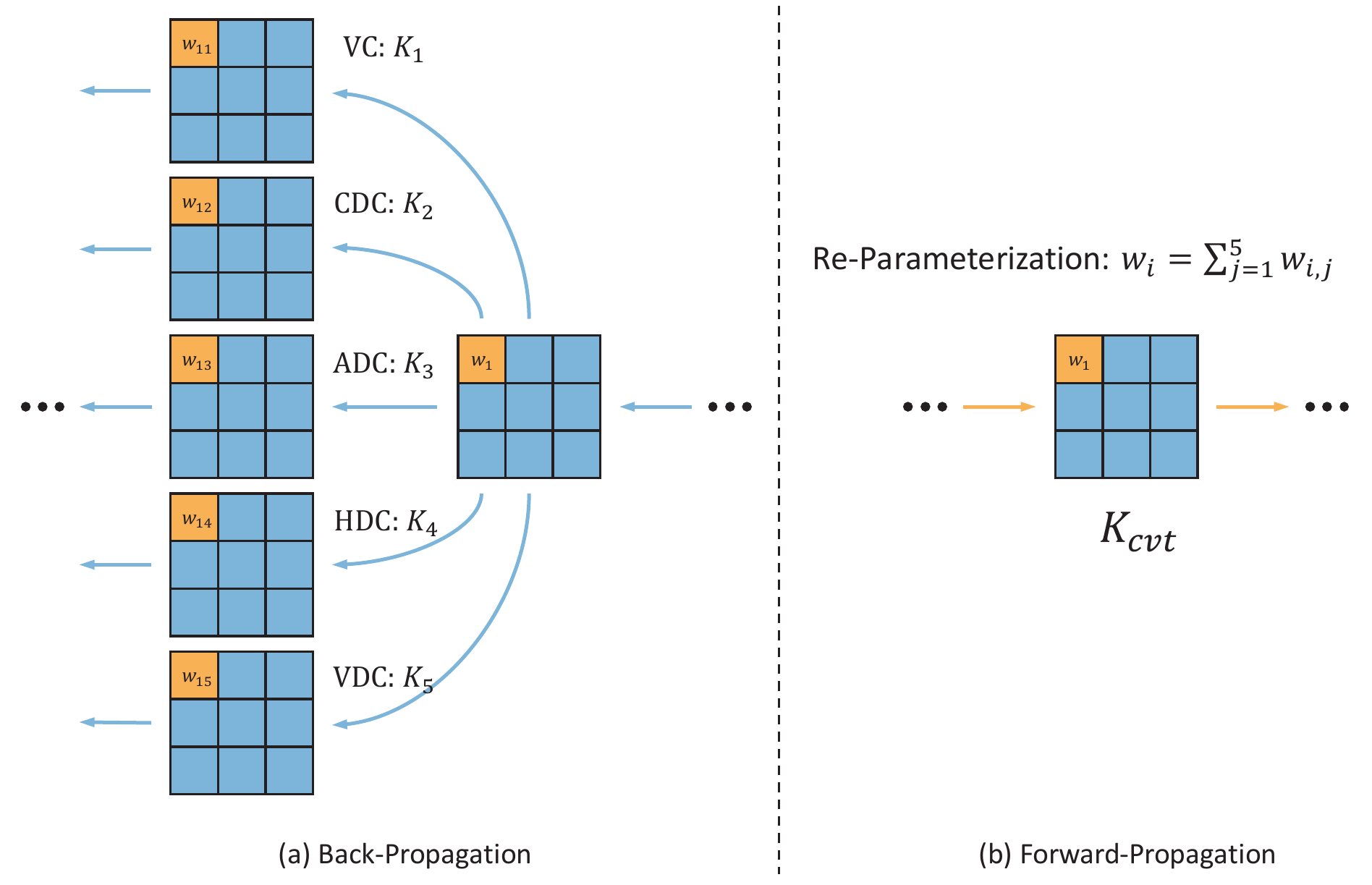}
	\caption{The process of the re-parameterization technique.}
	\label{fig:fig5}
\end{figure}

\begin{equation}
	\begin{aligned}
		F_{out} & = DEConv(F_{in}) = \sum\nolimits_{i=1}^{5}F_{in} \ast K_{i} \\
		&= F_{in} \ast (\sum\nolimits_{i=1}^{5}K_{i}) = F_{in} \ast K_{cvt},
	\end{aligned}
\end{equation}
where $DEConv(\cdot)$ denotes the operation of our proposed DEConv, $K_{i=1:5}$ represent the kernels of VC, CDC, ADC, HDC, and VDC, respectively, $\ast$ denotes the convolution operation, and $K_{cvt}$ denotes the converted kernel, which combines the parallel convolutions together.

Fig.~\ref{fig:fig5} visually shows the process of re-parameterization technique.
In the back-propagation phase, the kernel weights of five parallel convolutions are updated separately using the chain rule of gradient propagation.
In the forward-propagation phase, the kernel weights of the parallel convolutions are fixed and the converted kernel weights are calculated by adding up them on the corresponding positions.
Note that, the re-parameterization technique can accelerate the training and testing process simultaneously, since both of them contain the forward-propagation phase.


Compare with the vanilla convolution layer, the proposed DEConv can extract more richful features while maintains the parameter size, and introduces no extra computational cost and memory burdens in the inference stage.
More discussions about DEConv can be found in Sec.~\ref{subsec: DEConv}.

\subsection{Content-guided Attention}

Feature attention module (FAM) consists of a channel attention and a spatial attention, which are sequentially placed to calculate the attention weights in channel and spatial dimensions.
The channel attention calculates a channel-wise vector, i.e., $W_c \in \mathbb{R}^{C\times 1\times 1}$, to re-calibrate the features.
The spatial attention calculates a spatial importance map (SIM), i.e., $W_s \in \mathbb{R}^{H\times W}$ to adaptively indicate the importance levels of different regions.
The FAM treats different channels and pixels unequally, improving the dehazing performance.

\begin{figure}[!t]
	\centering
	\includegraphics[width=0.8\linewidth]{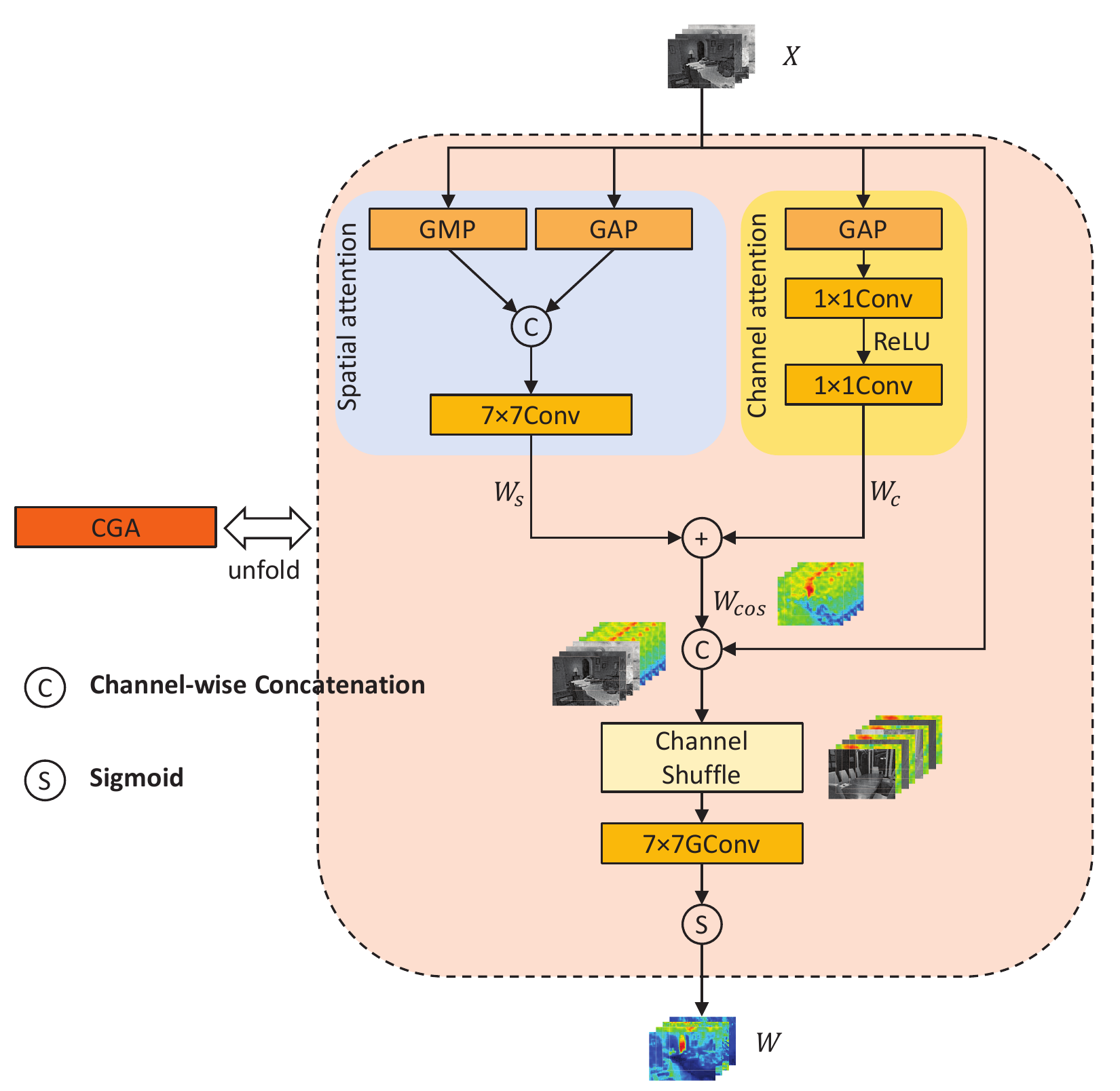}
	\caption{The diagram of content-guided attention (CGA). CGA is a coarse-to-fine process: the coarse version of SIMs (i.e., $W_{coa} \in \mathbb{R}^{C\times H\times W}$) is generated firstly and then every channel is refined by the guidance of input features.}
	\label{fig:fig6}
\end{figure}

However, the spatial attention inside FAM can only address the uneven haze distribution in image level, and ignore the uneven distribution in feature level.
The channel attention inside FAM models the channel-wise differences without considering the contextual information.
With the expanding of feature channels, the image-level haze distribution information is encoded into the feature maps.
Different channels in the feature space have different meanings depending on the role of the filters applied.
That means for every channel of features, the haze information is unevenly spread across the spatial dimensions.
The channel-specific SIMs are desired in this situation. 
In addition, another problem of FAM is that there is no information exchange between these two attention weights.
$W_c$ and $W_s$ are sequentially calculated and separately enhance the features.

To fully address the problems mentioned above, we propose a content-guided attention (CGA) to obtain the exclusive SIM for every single channel of input features in a coarse-to-fine manner, meanwhile fully mix channel attention weights and spatial attention weights to guarantee information interaction.
The detailed procedures of CGA are illustrated in Fig.~\ref{fig:fig6}, let $X \in \mathbb{R}^{C\times H\times W}$ denotes the proceeding input features, the goal of CGA is to generate channel-specific SIMs (i.e., $W \in \mathbb{R}^{C\times H\times W}$), which has the identical dimensions with $X$.

We first compute the corresponding $W_c$ and $W_s$ by following \cite{woo2018ECCV,Hu2018CVPR-SE}.
\begin{equation}
	\begin{aligned}
			W_c &= \mathcal{C}_{1\times 1}(max(0,\mathcal{C}_{1\times 1}(X_{GAP}^{c}))),\\
			W_s &= \mathcal{C}_{7\times 7}([X_{GAP}^{s},X_{GMP}^{s}]),
		\end{aligned}        
\end{equation}
where $max(0,x)$ denotes the ReLU activation function, $\mathcal{C}_{k\times k}(\cdot)$ denotes a convolution layer with $k\times k$ kernel size, $[\cdot]$ denotes the channel-wise concatenation operation.
$X_{GAP}^{c}$, $X_{GAP}^{s}$, and $X_{GMP}^{s}$ denote the features processed by global average pooling operation across the spatial dimensions, global average pooling operation across the channel dimension, and global max pooling operation across the channel dimension, respectively.
To reduce the number of parameters and limit the model complexity, the first $1\times 1$ convolution reduces the channel dimension from $C$ to $\frac{C}{r}$ ($r$ refers to the reduction ratio), and the second $1\times 1$ convolution expands it back to $C$.
In our implementation, we opt to reduce the channel dimension to a fixed value (i.e., 16) by setting $r$ to $\frac{C}{16}$.

Then we fuse $W_c$ and $W_s$ together via a simple addition operation, which follows broadcasting rules, to obtain the coarse SIMs $W_{coa} \in \mathbb{R}^{C\times H\times W}$.
We experimentally find the product operation can achieve similar results.
\begin{equation}
	W_{coa} = W_c + W_s,
\end{equation}

In order to obtain the final refined SIMs $W$, every channel of $W_{coa}$ is adjusted according to corresponding input features.
We utilize the content of input features as the guidance to generate the final channel-specific SIMs $W$.
In particular, every channel of $W_{coa}$ and $X$ are re-arranged in an alternating manner via a channel shuffle operation \cite{Zhang2018CVPR-shufflenet}.

\begin{equation}
	W = \sigma(\mathcal{GC}_{7\times 7}(CS([X,W_{coa}]))),
\end{equation}
where $\sigma$ denotes the sigmoid operation, $CS(\cdot)$ denotes the channel shuffle operation, $\mathcal{GC}_{k\times k}(\cdot)$ denotes a group convolution layer with $k\times k$ kernel size, and in our implementation, the group number is set to $C$.

The CGA assigns unique SIM to every channel, guiding the model to focus on significant regions of each channel.
Therefore, more useful information encoded in features can be emphasized to effectively improve the dehazing performance.

As shown in the right part of Fig.~\ref{fig:overall structure},
combining proposed DEConv with the CGA, we propose the main block of our DEA-Net, i.e., detail-enhanced attention block (DEAB).
By removing the CGA part, we obtain the detail enhanced block (DEB).

\subsection{CGA-based Mixup Fusion Scheme}
Following \cite{dong2020CVPR,bai2022TIP,wu2021CVPR,hong2022AAAI}, we adopt the encoder-decoder-like (or U-Net-like) architecture for our DEA-Net.
We observe that fusing the features from the encoder part with that from the decoder part is an effective trick in dehazing and other low-level vision tasks \cite{He2021SP-ESKN,He2023PR,dong2020CVPR,wu2021CVPR}.
Low-level features (e.g., edges and contours), which have a non-negligible role for recovering haze-free images, gradually lose their impact after passing through many intermediate layers.
Feature fusion can enhance the information flow from shallow layers to deep ones, which is beneficial for feature preserving and gradient back-propagation.
The simplest way for fusion is element-wise addition, which is adopted in many previous approaches \cite{dong2020CVPR,bai2022TIP,ye2022ECCVORAL}.
Later, Wu et al. \cite{wu2021CVPR} applied the adaptive mixup operation to adjust the fusion proportion via self-learned weights, which is more flexible than the addition.

However, there exists a receptive field mismatch problem in above mentioned fusion schemes.
The information encoded in the shallow features is tremendously different from the information encoded in the deep features, since they have the totally different receptive fields. 
One single pixel in the deep features are originated from a region of pixels in the shallow features.
Simple addition or concatenation operation or mixup operation fails to address the mismatch before fusion.

To mitigate this problem, we further propose a CGA-based mixup scheme to adaptively fuse the low-level features in the encoder part with corresponding high-level features, by modulating the features via learned spatial weights.


Fig.~\ref{fig:overall structure}~(d) shows the details of proposed CGA-based mixup fusion scheme.
The core part is that we opt to employ the CGA to calculate the spatial weights for feature modulation.
The low-level features in the encoder part and corresponding high-level features are fed into the CGA to calculate the weights, and then combined by a weighted summation method.
We also add the input features via skip connections to mitigate gradient vanishing problem and ease the learning process.
Finally, the fused features are projected by a $1\times 1$ convolution layer to obtain the final features (i.e., $F_{fuse}$). 
\begin{equation}
	F_{fuse} = \mathcal{C}_{1\times 1}(F_{low}\cdot W + F_{high} \cdot (1-W) + F_{low} + F_{high}),
\end{equation}

More discussions about the CGA-based mixup fusion scheme can be found in Sec.~\ref{subsec: CGA-based Mixup Fusion Scheme}.

\subsection{Overall Architecture}
By combining (1) DEConv, (2) CGA, and (3) CGA-based mixup fusion scheme together, we propose our DEA-Net with DEAB and DEB as the basic blocks.
As shown in Figure \ref{fig:overall structure}, our DEA-Net is a three-level encoder-decoder-like (or U-Net-like) architecture, which consists of three parts: encoder part, feature transform part, and decoder part.
There are two down-sampling operations and two up-sampling operations in our DEA-Net.
The down-sampling operation halves the spatial dimensions and doubles the number of channels. 
It is realized through a normal convolution layer by setting the value of stride to 2 and setting the number of output channels to 2 times of input channels.
The up-sampling operation can be regarded as the inverse form of the down-sampling operation, which is realized through a deconvolution layer.
The dimensional size of level 1, level 2, and level 3 are $C\times H\times W$, $2C\times \frac{H}{2} \times \frac{W}{2}$, and $4C\times \frac{H}{4} \times \frac{W}{4}$, respectively.
In our implementation, we set the value of $C$ to 32.
Previous methods \cite{wu2021CVPR,hong2022AAAI} transform the features only in the low-resolution space, resulting in information loss, which is non-trivial for the detail-sensitive task like dehazing. 
Differently, we deploy feature extraction blocks from level 1 to level 3.
Specifically, we opt to employ different blocks in different levels (level 1\&2: DEB, level 3: DEAB).
For feature fusion, we fuse the features after the down-sampling operations and corresponding features before the up-sampling operations (highlighted with green arrow lines in Fig.~\ref{fig:overall structure}).
Finally, we simply employ a $3 \times 3$ convolution layer at the end to obtain the dehazing result $J$.


The DEA-Net is trained by minimizing the pixel-wise difference between the predicted haze-free image $J$ and the corresponding ground truth $GT$.
In our implementation, we choose $L1$ loss function (i.e., mean absolute error) to drive the training.
\begin{equation}
	\mathcal{L}_{L1} = ||J-GT||_1,
\end{equation}

\section{Experiment}
\label{sec: experiment}

\subsection{Datasets and Metrics}
\textbf{Datasets.} In our implementation, we train and test our proposed DEA-Net on synthetic and real-captured datasets. 
REalistic Single Image DEhazing (RESIDE) \cite{li2018TIP} is a widely-used dataset, which contains five subsets: Indoor Training Set (ITS), Outdoor Training Set (OTS), Synthetic Objective Testing Set (SOTS), Real-world Task-driven Testing Set (RTTS), and Hybrid Subjective Testing Set (HSTS).
We select ITS and OTS in the training phase and select SOTS in the testing phase.
Note that, the SOTS is divided into two subsets (i.e., SOTS-indoor and SOTS-outdoor) for evaluating the models separately trained on ITS and OTS.
ITS contains 1399 indoor clean images and for every clean image, 10 simulated hazy images are generated based on the physical scattering model.
As for OTS, we pick around 296K images for the training process \footnote{Following \cite{liu2019ICCV}, data cleaning is applied since the intersection of training and testing datasets.}.
SOTS-indoor and SOTS-outdoor contain 500 indoor and 500 outdoor testing images, respectively.
In addition, Haze4K dataset \cite{liu2021ACMMM}, which contains 3000 synthetic training images and 1000 synthetic testing images, is also employed to evaluate our DEA-Net.
Besides, some real-captured hazy images are utilized to further verify the effectiveness on real scenes.

\textbf{Evaluation Metrics.} Peak signal-to-noise-ratio (PSNR) and structural similarity index (SSIM) \cite{Wang2004TIP-SSIM}, which are commonly used to measure the image quality among the computer vision community, are utilized for dehazing performance evaluation.
For a fair comparison, we calculate the metrics based on the RGB color images without cropping pixels.

\subsection{Implementation Details}
We implement the proposed DEA-Net model on PyTorch deep learning platform with a single NVIDIA RTX3080Ti GPU.
We deploy DEB, DEB, and DEAB in level 1, level 2, and level 3, respectively.
The number of blocks deployed on different stages $[N_1, N_2, N_3, N_4, N_5]$ is set to $[4, 4, 8, 4, 4]$.
The DEA-Net is optimized using Adam \cite{Kingma2015ICLR-Adam} optimizer and $\beta_1$, $\beta_2$, $\varepsilon$ are set to default values, i.e., $0.9$, $0.999$, $1e^{-8}$.
Moreover, the initial learning rate and the batch size are set to $1e^{-4}$ and $16$, respectively.
Cosine annealing strategy \cite{He2019CVPR-Bag} is adopted to adjust the learning rate from the initial value to $1e^{-6}$.
To train the model, we randomly crop patches from the original images with size $256\times 256$, then 
two data augmentation techniques are adopted including: $90^{\circ}$ or $180^{\circ}$ or $270^{\circ}$ rotation and vertical or horizontal flip.
In the whole training phase, the model is trained for 1500K iterations, and it takes roughly 5 days to train our DEA-Net on ITS.

\subsection{Ablation Study}
To demonstrate the effectiveness of our proposed DEA-Net, we investigate on the designs and effects of (1) Detail-enhanced convolution (DEConv), (2) Content-guided attention (CGA), and (3) CGA-based mixup fusion scheme.
The contribution of each components are analyzed via ablation experiments.

\subsubsection[1]{DEConv}
\label{subsec: DEConv}
We first construct the baseline model by deploying classical residual block (RB) \cite{he2016deep} in level 3, and this model is denoted as \textit{Base\_RB}. 
As a popular basic block used in dehazing domain, we also employ the feature attention block (FAB) from \cite{qin2020AAAI} in level 3.
The hyper-parameters are set to the default values as described in the original paper.
We term this model as our second baseline, \textit{Base\_FAB}.

\begin{figure}[h]
	\centering
	\includegraphics[width=1\linewidth]{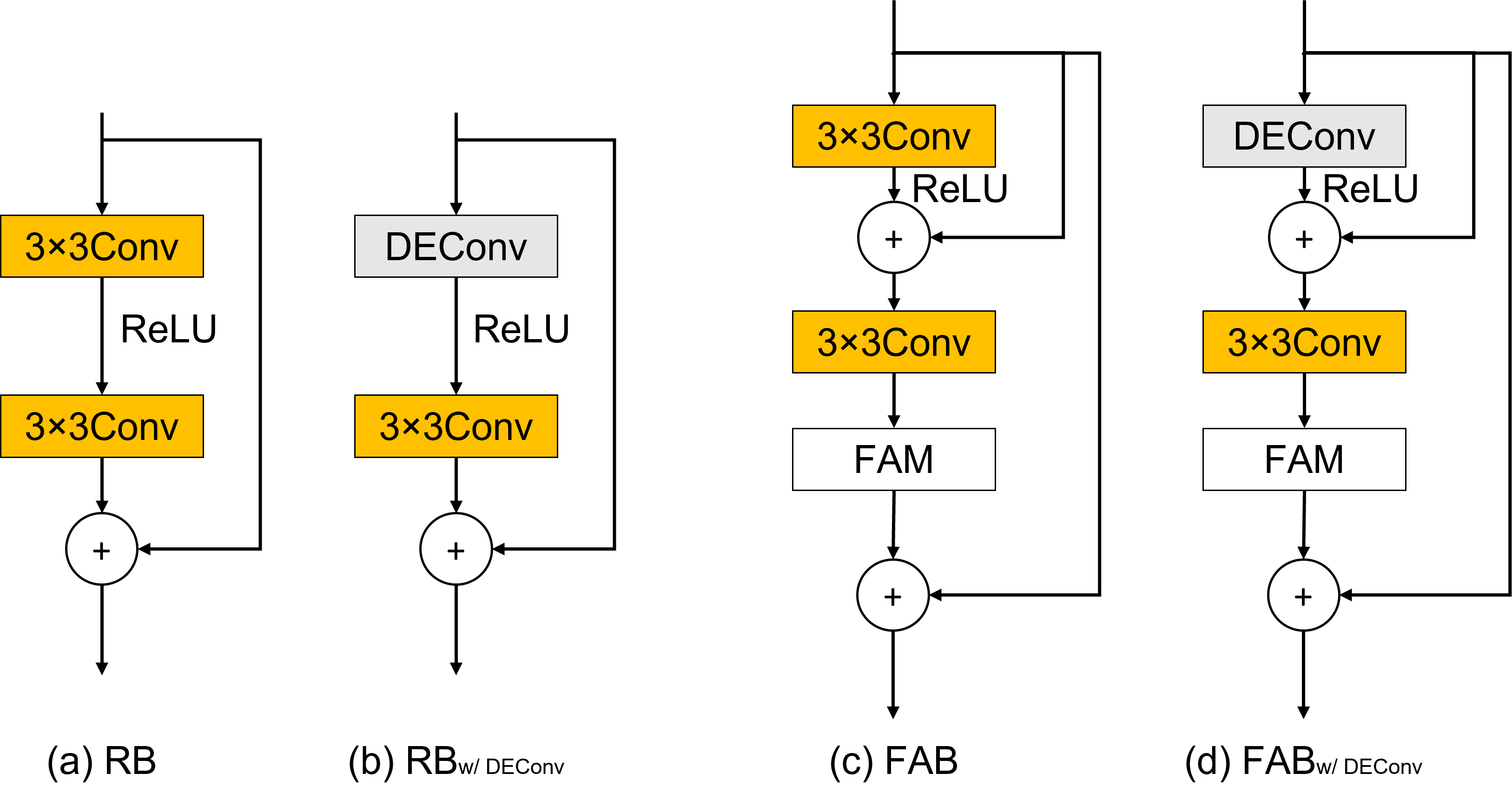}
	\caption{The schematic diagrams of RB, RB{\tiny w/ DEConv}, FAB, and FAB{\tiny w/ DEConv}. The first vanilla convolution layer in RB/FAB is replaced with the proposed DEConv to generate the RB{\tiny w/ DEConv}/FAB{\tiny w/ DEConv}.}
	\label{fig:figX}
\end{figure}

To extract more effective features, we revise the block by introducing DEConv into RB and FAB.
As illustrated in Fig.~\ref{fig:figX}, the first vanilla convolution layer is replaced with the proposed DEConv in both RB and FAB.
The blocks deployed in level 3 are indicated as RB{\tiny w/ DEConv} and FAB{\tiny w/ DEConv}, respectively.
The corresponding models are denoted as \textit{Model\_RB\_D} and \textit{Model\_FAB\_D}.

\begin{table*}[h]
	\footnotesize
	\centering
	\caption{Ablation study of DEConv and CGA. All the experiments are conducted on SOTS-Indoor \cite{li2018TIP} dataset.}
	\label{tab:table1}
	\begin{tabular}{c|c|cccccc}
		\toprule
		\multicolumn{2}{c|}{Model}         & Base\_RB     & Base\_FAB     & Model\_RB\_D & Model\_FAB\_D & Model\_DEAB & Model\_FAB\_D\_CBAM  \\
		\midrule \midrule
		\multirow{4}{*}{Setting} & Level 1 & --    & --    & -- & -- & --&-- \\
		& Level 2 & --  & --  & -- &-- &-- &--\\
		& Level 3 & RB & FAB & RB{\tiny w/ DEConv} & FAB{\tiny w/ DEConv} & FAB{\tiny w/ DEConv \& CGA} (DEAB) & FAB{\tiny w/ DEConv \& CBAM} \\ 
		&Attention & -- & FAM & -- & FAM  & CGA & CBAM \cite{woo2018ECCV} \\ 
		\midrule \midrule
		\multicolumn{2}{c|}{PSNR (dB)}     & 30.74      &  33.07     & 31.01   &  33.67  & 35.17 & 34.68 \\
		\multicolumn{2}{c|}{SSIM}          & 0.9729      &  0.9824     & 0.9739  &  0.9840  & 0.9866 & 0.9857\\
		\multicolumn{2}{c|}{\# Param. (K)}  & 2105      & 2143    & 4467   & 4505    & 4569& 4493 \\ 
		\bottomrule
	\end{tabular}
\end{table*}

For a fair comparison, all of the four blocks (i.e., RB, FAB, RB{\tiny w/ DEConv}, and FAB{\tiny w/ DEConv}) are cascaded for 6 times in level 3, and the same fusion scheme is used (i.e., Mixup \cite{qin2020AAAI}).
For convenience, we omit the blocks in level 1 and level 2, and train the models for only 500K iterations with initial learning rate is set to $2e^{-4}$ (These settings are with the ablation study).
The experimental results are tested on the same testing dataset (i.e., SOTS-Indoor \cite{li2018TIP} dataset).
Although metrics are lower than the completely trained models reported in Table.~\ref{tab:benchmark}, the trends and values are consistent and meaningful.

The performance of all these aforementioned models is summarized in Table.~\ref{tab:table1}.
Replacing the vanilla convolution layer with the parallel convolution layers (i.e., DEConv) brings 0.27 and 0.6 dB improvement in terms of PSNR on RB and FAB, respectively.
By comparing \textit{Model\_FAB\_D} with \textit{Base\_FAB}, the results indicate that DEConv can definitely improve the values of metrics (i.e., PSNR and SSIM) at the cost of around twice number of parameters (4505 K \emph{vs.} 2143 K).
That is very unfriendly and may cause failure under some memory-limited situations, prohibiting the usage of the DEConv on mobile or embedded devices.

In order to deal with the problem, we equivalently transform the DEConv into a standard $3\times 3$ convolution by adding up the learned kernel weights in the same positions (i.e., re-parameterization).
Table.~\ref{tab:re-param} shows the comparative results of the number of parameters (\# Param.), the number of floating-point operations (\# FLOPs) and inference time of \textit{Model\_FAB\_D} before and after the re-parameterization operation.
We can clearly see that the re-parameterization operation simplifies the parallel structure without triggering performance drop.
In particular, after the simplification, \textit{Model\_FAB\_D} still achieves 0.6 dB performance improvement when comparing with \textit{Base\_FAB} and no extra overhead is introduced. 

\begin{table}[!t]
	\footnotesize
	\centering
	\caption{The comparative results of the number of parameters (\# Parameters), the number of floating-point operations (\# FLOPs) and inference time of \textit{Model\_FAB\_D} before and after the re-parameterization operation. Re-Pa. is short for the re-parameterization operation.}
	\label{tab:re-param}	
	\begin{tabular}{c|cc|c}
		\toprule
		 & \multicolumn{2}{c|}{Model\_FAB\_D}& Base\_FAB\\
		 & w/o Re-Pa.  & w/ Re-Pa. & -- \\
		\midrule
		\midrule
		\# Param. (K) & 4505 & 2143 & 2143  \\
		\# FLOPs (G)  & 23.72 & 9.23 & 9.23 \\
		inference time (ms) &4.53  & 1.76 & 1.76 \\
		PSNR (dB) & 33.67 & 33.67 & 33.07\\
		\bottomrule
	\end{tabular}	
\end{table}

In addition, we also explore the designs of parallel convolution layers from only a single vanilla convolution layer (i.e., FAB) to two parallel vanilla convolution layers, then to the complete DEConv (i.e., FAB{\tiny w/ DEConv}).
As shown in Table.~\ref{tab:deconv}, adding a parallel vanilla convolution layer to the FAB causes a 0.15 dB performance drop.
The underlying reason behind this may be the training difficulty due to redundant features extracted by the identical layers.
On the contrary, adding a parallel CDC layer to the FAB boosts the performance.
The experimental results verify that by embedding traditional prior information, difference convolution (DC) layers can effectively extract more representative features. 
We also observe that by adding more parallel DC streams for feature extraction, the performance gradually improves from 33.07 dB to 33.67 dB in terms of PSNR. Similar trend can be observed in terms of SSIM.
Based on the discussions above, we choose \textit{Model\_FAB\_D} with basic block FAB{\tiny w/ DEConv} for the following study.

\begin{table}[ht]
	\footnotesize
	\centering
	\caption{The experimental results on designs of parallel convolution layers. \CheckmarkBold \CheckmarkBold means the same convolution layer is used twice within two parallel streams. The metrics are tested on SOTS-Indoor \cite{li2018TIP} dataset.}
	\label{tab:deconv}	
	\begin{tabular}{l|ccccc}
		\toprule
		Design &FAB &$+$ vanilla &$+$ DC&$+$ DC &FAB{\tiny w/ DEConv} \\
		\midrule
		\midrule
		Vanilla Conv. & \CheckmarkBold &\CheckmarkBold \CheckmarkBold & \CheckmarkBold & \CheckmarkBold & \CheckmarkBold \\
		$+$ CDC & && \CheckmarkBold & \CheckmarkBold & \CheckmarkBold \\
		$+$ HDC \& VDC & && & \CheckmarkBold & \CheckmarkBold \\
		$+$ ADC & && & & \CheckmarkBold \\
		\midrule
		PSNR & 33.07 & 32.92& 33.23 & 33.43 & \textbf{33.67} \\
		SSIM & 0.9824 & 0.9820 & 0.9826 & 0.9833 & \textbf{0.9840} \\
		\bottomrule
	\end{tabular}	
\end{table}

\subsubsection[2]{CGA}
Further, we investigate the effectiveness of the proposed two-step coarse-to-fine attention mechanism (i.e., CGA).
As mentioned in Section \ref{sec: introduction}, CGA generates channel-specific spatial importance maps (SIMs) to indicate the important regions of individual channel.
We compare the CGA with the other attention mechanisms such as feature attention module (FAM) used in many dehazing methods \cite{qin2020AAAI,wu2021CVPR,Wu2020CVPRW} and the common convolutional block attention module (CBAM) \cite{woo2018ECCV}.
Both FAM and CBAM contain sequential channel attention and spatial attention with slightly different implementations.

\textit{Model\_FAB\_D} cascades FAB{\tiny w/ DEConv} blocks in level 3 and inside the FAB{\tiny w/ DEConv} block, the FAM is adopted.
Then, we combine CGA and CBAM into FAB{\tiny w/ DEConv} block to generate the FAB{\tiny w/ DEConv \& CGA} (i.e., DEAB) and FAB{\tiny w/ DEConv \& CBAM}, respectively, and the corresponding models are denoted as \textit{Model\_DEAB} and \textit{Model\_FAB\_D\_CBAM}.

The spatial attention used in FAM or CBAM learns the SIM with only one single channel to indicate the important regions of the input features with relatively larger number of channels.
Such approaches neglect the specificity of each channel of features and somehow restrict the powerful representation ability of CNNs.
As shown in the right three columns of Table.~\ref{tab:table1}, \textit{Model\_DEAB} outperforms both \textit{Model\_FAB\_D} and \textit{Model\_FAB\_D\_CBAM} by 1.5 dB and 1.01 dB in terms of PSNR.
The results indicate that CGA can better re-calibrate the features via learning channel-specific SIMs to pay attention to channel-wise haze distribution difference.

\begin{figure}[h]
	\centering
	\subfloat[Hazy]{\includegraphics[width=0.2\textwidth]{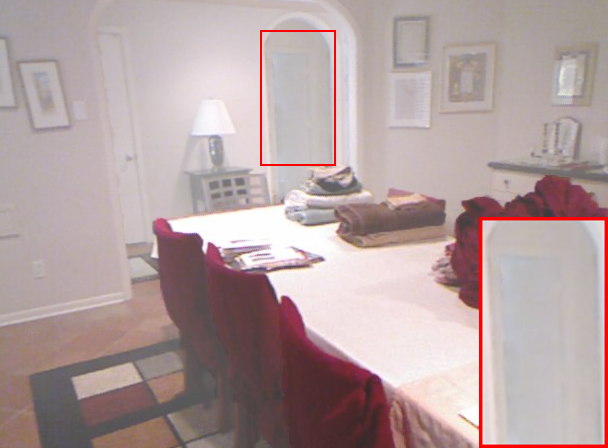}}
	\hfil
	\subfloat[GT]{\includegraphics[width=0.2\textwidth]{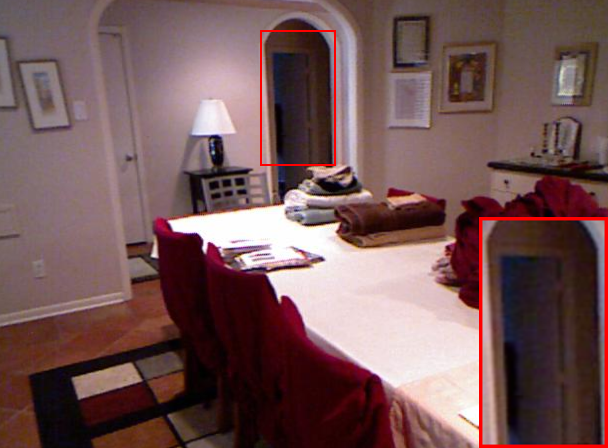}}
	\vspace{-3mm}
	\subfloat[Model\_FAB\_D]{\includegraphics[width=0.2\textwidth]{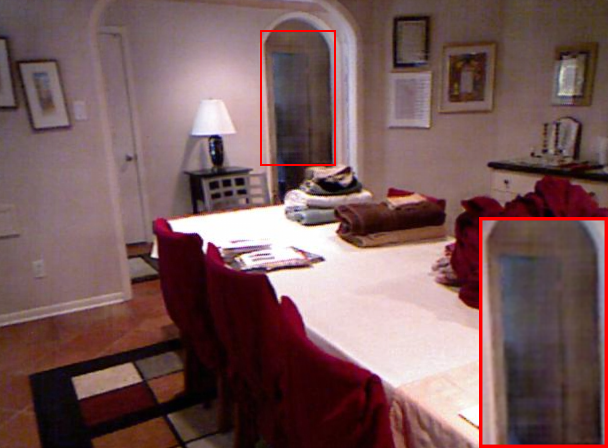}%
		\label{fig:CGA_FA_output}}
	\hfil
	\subfloat[Model\_DEAB]{\includegraphics[width=0.2\textwidth]{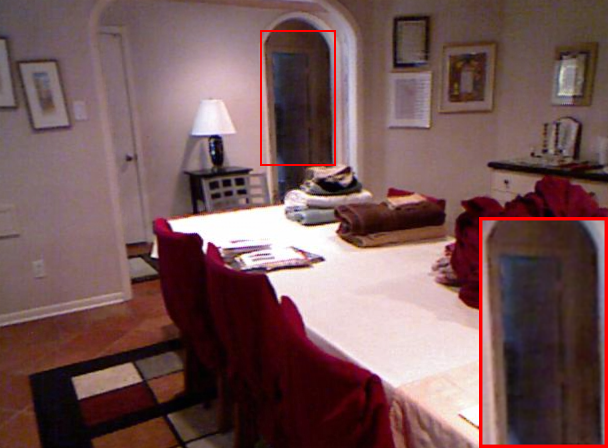}%
		\label{fig:CGA_CGA_outpu}}
	\vspace{-3mm}
	\subfloat[SIM of (c)]{\includegraphics[width=0.2\textwidth]{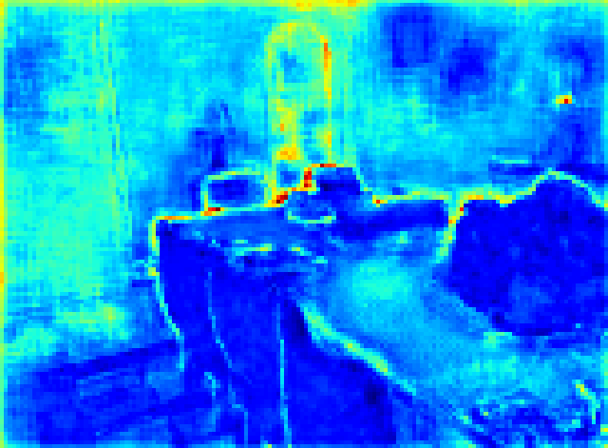}%
		\label{fig:CGA_FA_attn}}
	\hfil
	\subfloat[SIMs of (d)]{\includegraphics[width=0.2\textwidth]{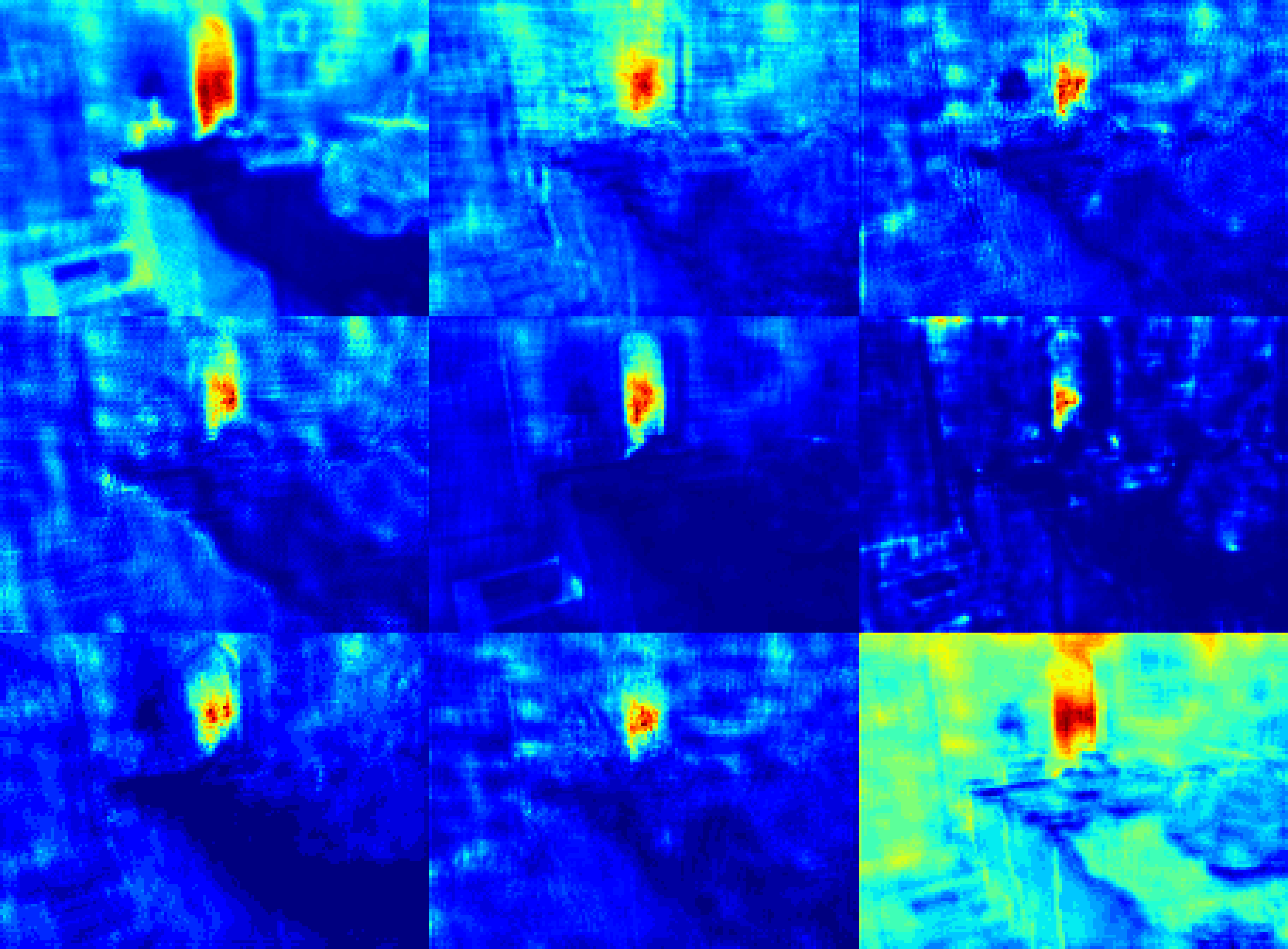}%
		\label{fig:CGA_CGA_attn}}
	\caption{Visual comparisons of FAM and our proposed CGA. We show the learned SIMs and corresponding results.}
	\label{fig:CGA_attn}
\end{figure}

Fig.~\ref{fig:CGA_attn} visually illustrates the SIMs learned by CGA and FAM and the corresponding processing results.
As we can see from Fig.~\ref{fig:CGA_FA_attn}, one-channel SIM obtained by FAM can indicate the uneven haze distribution (to some extend).
However, it is not accurate enough (e.g., the red chairs region) due to the mix of some contour patterns.
By using the content of input features to guide the generation of SIMs, CGA can learn more accurate spatial weights.
Fig.~\ref{fig:CGA_CGA_attn} shows eight randomly selected channels of SIMs, and the average map of all SIMs (right bottom).
The channel-specific SIMs treat different channels of features with different spatial weights, which can better guide the model to focus on critical regions.
Fig.~\ref{fig:CGA_FA_output} and Fig.~\ref{fig:CGA_CGA_outpu} are the corresponding results.
We observe that the arched door region (highlighted by the red rectangle) recovered by \textit{Model\_FAB\_D} has obvious haze residual.

\begin{table*}[p]
	\footnotesize
	\centering
	\caption{Ablation study of CGA-based mixup fusion scheme. We compare it with element-wise addition and mixup \cite{wu2021CVPR}. All the experiments are conducted on SOTS-Indoor \cite{li2018TIP} dataset.}
	\label{tab:table4}
	\begin{tabular}{c|c|ccc|cc}
		\toprule
		\multicolumn{2}{c|}{Model}         &  Model\_DEAB\_A & Model\_DEAB & Model\_DEAB\_C & Model\_MS & DEA-Net-S   \\ 
		\midrule \midrule
		\multirow{4}{*}{Setting} & Level 1 & --    & --    & -- & RB&DEB \\
		& Level 2 & --  & --  & -- &RB &DEB \\
		& Level 3 &DEAB & DEAB & DEAB & DEAB&DEAB \\
		&Fusion scheme & Addition & Mixup & CGA-based Mixup& CGA-based Mixup&CGA-based Mixup\\ 
		\midrule \midrule
		\multicolumn{2}{c|}{PSNR (dB)}     & 35.23  & 35.17  & 35.40 & 37.92 &39.16\\
		\multicolumn{2}{c|}{SSIM}          & 0.9864 & 0.9866  & 0.9875 & 0.9915&0.9921 \\
		\bottomrule
	\end{tabular}
\end{table*}

\begin{table*}[!t]
	\footnotesize
	\centering
	\caption{Quantitative comparisons of various dehazing methods on SOTS-indoor, SOTS-ourdoor, and Haze4K. We report PSNR, SSIM, number of parameters (\# Param.), number of floating-point operations (\# FLOPs), and runtime to perform comprehensive comparisons. The sign ``-'' denotes the digit is unavailable. \textbf{Bold} and \underline{underlined} indicate the best and the second best results, respectively.}
	\label{tab:benchmark}
	\begin{tabular}{l|cc|cc|cc|ccc}
		\toprule
		\multirow{2}{*}{Method} & \multicolumn{2}{c|}{SOTS-indoor \cite{li2018TIP}} & \multicolumn{2}{c|}{SOTS-outdoor \cite{li2018TIP}} & \multicolumn{2}{c|}{Haze4K \cite{liu2021ACMMM}} & \multicolumn{3}{c}{Overhead}\\
		& PSNR & SSIM & PSNR & SSIM & PSNR & SSIM & \# Param. (M) & \# FLOPs (G) & Runtime (ms) \\
		\midrule
		\midrule
		(TPAMI'10) DCP \cite{he2010TPAMI} & 16.61 & 0.8546 & 19.14 & 0.8605 & 14.01 & 0.76 & - & - & - \\
		(TIP'16) DehazeNet \cite{cai2016TIP} & 19.82 & 0.8209 & 27.75 & 0.9269 & 19.12 & 0.84 & 0.008 & 0.5409 & 0.9932 \\
		(ICCV'17) AOD-Net \cite{li2017ICCV} & 20.51 & 0.8162 & 24.14 & 0.9198 & 17.15 & 0.83 & 0.0018 & 0.1146 & 0.3159 \\
		(CVPR'18) GFN \cite{ren2018CVPR} & 22.30 & 0.8800 & 21.55 & 0.8444 & - & - & 0.4990 & 14.94 & - \\
		\midrule
		(AAAI'20) FFA-Net \cite{qin2020AAAI} & 36.39 & 0.9886 & 33.57 & 0.9840 & 26.97 & 0.95 & 4.456 & 287.5 & 47.98 \\
		(CVPR'20) MSBDN \cite{dong2020CVPR} & 32.77 & 0.9812 & 34.81 & 0.9857 & 22.99 & 0.85 & 31.35 & \textbf{24.44} & 9.826 \\
		(ACMMM'21) DMT-Net \cite{liu2021ACMMM} & - & - & - & - & 28.53 & 0.96 & 51.79 & 75.56 & 26.83 \\
		(CVPR'21) AECR-Net \cite{wu2021CVPR} & 37.17 & 0.9901 & - & - & - & - & \textbf{2.611} & 52.20 & - \\
		(TIP'22) SGID-PFF \cite{bai2022TIP} & 38.52 & 0.9913 & 30.20 & 0.9754 & - & - & 13.87 & 152.8 & 20.92 \\
		(AAAI'22) UDN \cite{hong2022AAAI} & 38.62 & 0.9909 & 34.92 & 0.9871 & - & - & 4.250 & - & - \\
		(ECCV'22) PMDNet \cite{ye2022ECCVORAL} & 38.41 & 0.9900 & 34.74 & 0.9850 & \underline{33.49} & \underline{0.98} & 18.90 & - & - \\
		(CVPR'22) Dehamer \cite{guo2022CVPR} & 36.63 & 0.9881 & 35.18 & 0.9860 & - & - & 132.4 & 48.93 & 14.12 \\
		\midrule
		(Ours) DEA-Net-S & 39.16&0.9921&-&-&-&-&\underline{2.844}&\underline{24.88}&\textbf{5.632}\\	
		(Ours) DEA-Net & \underline{40.20} & \underline{0.9934} & \underline{36.03} & \underline{0.9891} & 33.19 & \textbf{0.99} & 3.653 & 32.23 & \underline{7.093} \\
		(Ours) DEA-Net-CR & \textbf{41.31} & \textbf{0.9945} & \textbf{36.59} & \textbf{0.9897} & \textbf{34.25} & \textbf{0.99} & 3.653 & 32.23 & \underline{7.093} \\
		\bottomrule
	\end{tabular}
\end{table*}
\begin{figure*}[p]
	\centering
	\subfloat[Hazy]{
		\begin{minipage}[c]{0.12\textwidth} 
			\includegraphics[width=1\textwidth]{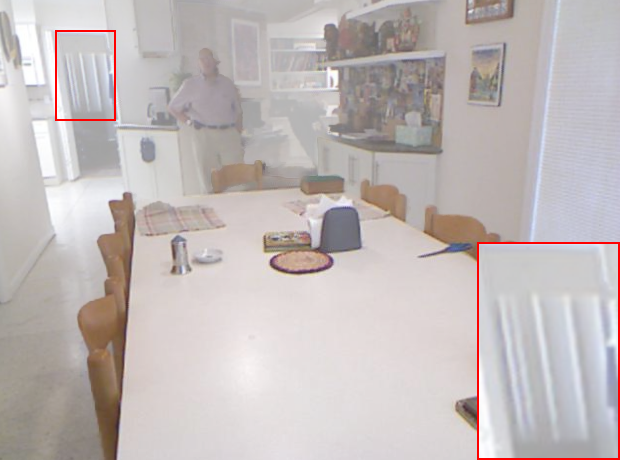}\vspace{2pt}
			\includegraphics[width=1\textwidth]{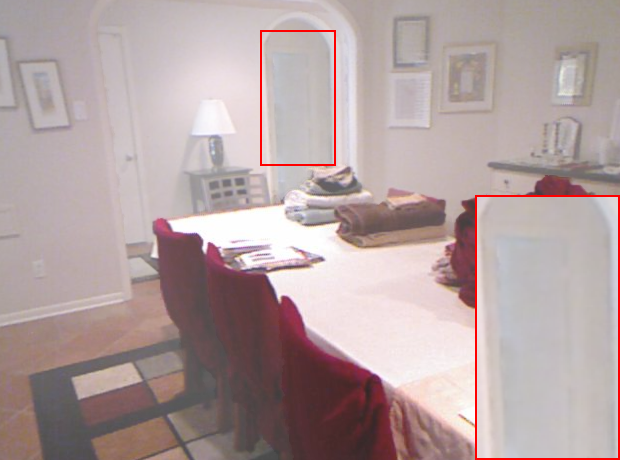}\vspace{2pt}
		\end{minipage}
	}	
	\hspace{-9pt}
	\subfloat[DCP \cite{he2010TPAMI}]{
		\begin{minipage}[c]{0.12\textwidth} 
			\includegraphics[width=1\textwidth]{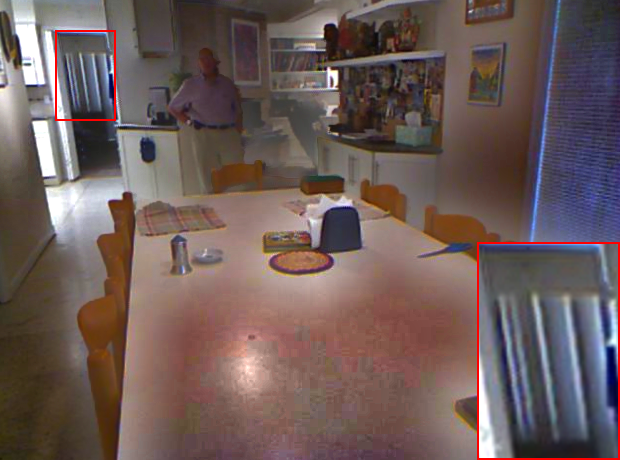}\vspace{2pt}
			\includegraphics[width=1\textwidth]{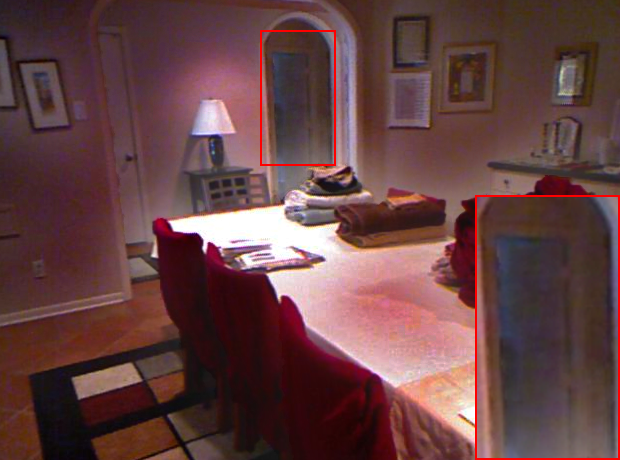}\vspace{2pt}
		\end{minipage}
	}	
	\hspace{-9pt}
	\subfloat[GDN \cite{liu2019ICCV}]{
		\begin{minipage}[c]{0.12\textwidth} 
			\includegraphics[width=1\textwidth]{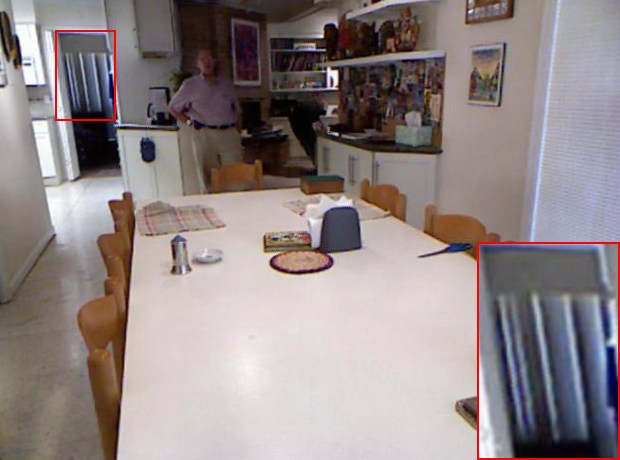}\vspace{2pt}
			\includegraphics[width=1\textwidth]{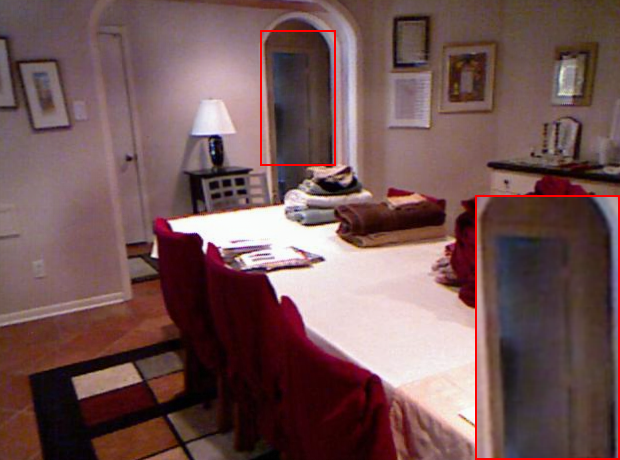}\vspace{2pt}
		\end{minipage}
	}	
	\hspace{-9pt}
	\subfloat[FFA \cite{qin2020AAAI}]{
		\begin{minipage}[c]{0.12\textwidth} 
			\includegraphics[width=1\textwidth]{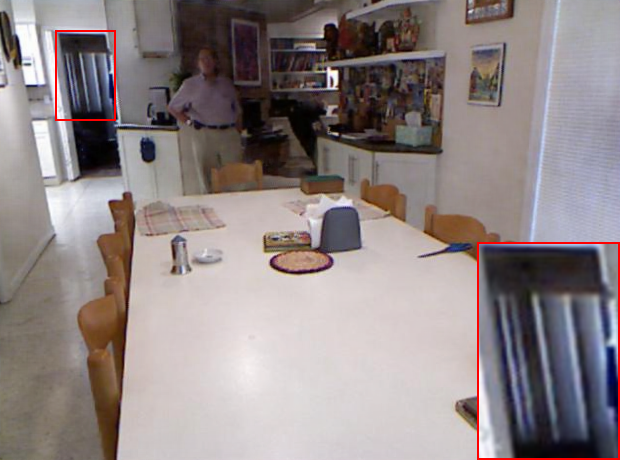}\vspace{2pt}
			\includegraphics[width=1\textwidth]{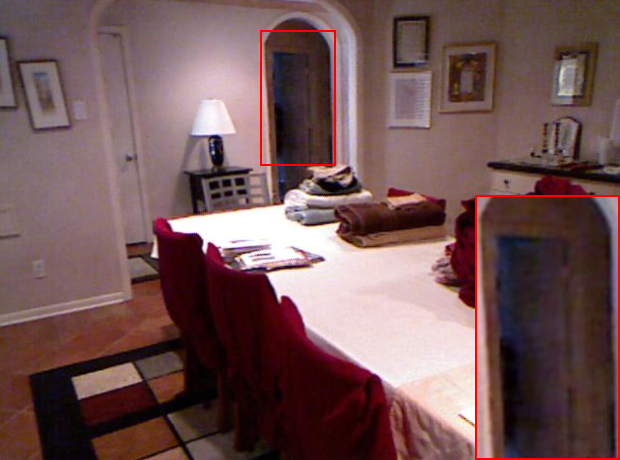}\vspace{2pt}
		\end{minipage}
	}	
	\hspace{-9pt}
	\subfloat[AECR-Net \cite{wu2021CVPR}]{
		\begin{minipage}[c]{0.12\textwidth} 
			\includegraphics[width=1\textwidth]{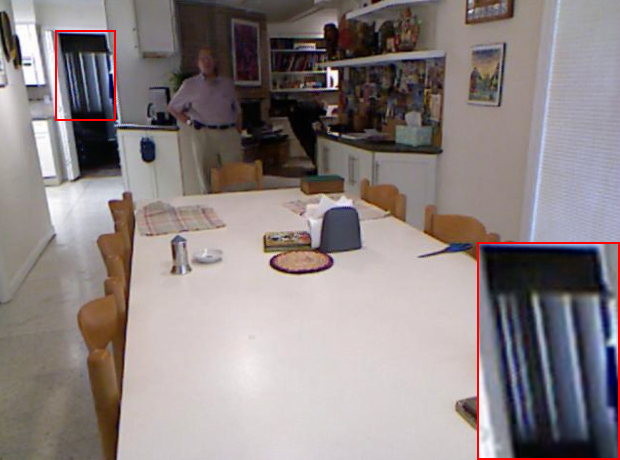}\vspace{2pt}
			\includegraphics[width=1\textwidth]{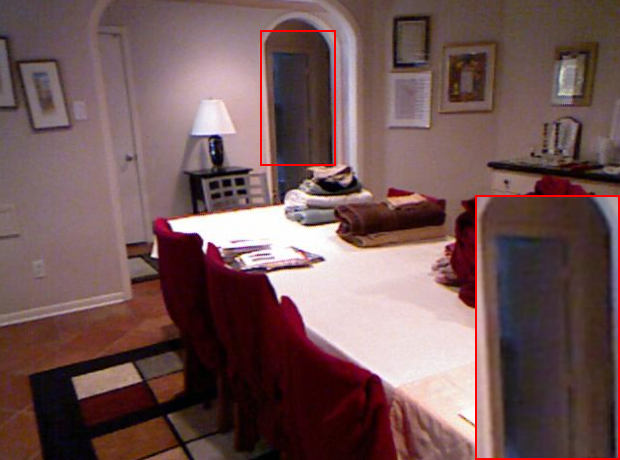}\vspace{2pt}
		\end{minipage}
	}	
	\hspace{-9pt}
	\subfloat[Ours]{
		\begin{minipage}[c]{0.12\textwidth} 
			\includegraphics[width=1\textwidth]{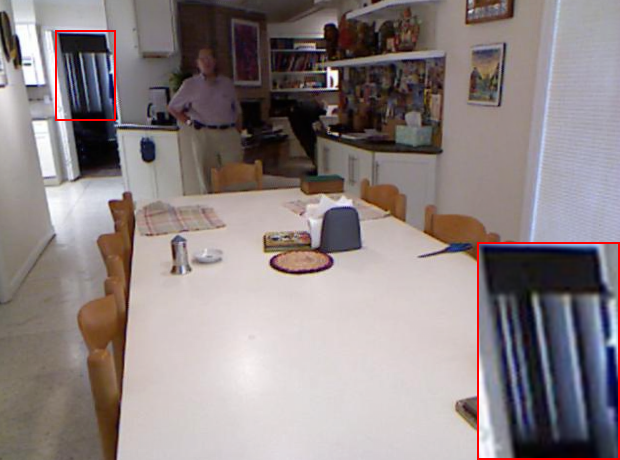}\vspace{2pt}
			\includegraphics[width=1\textwidth]{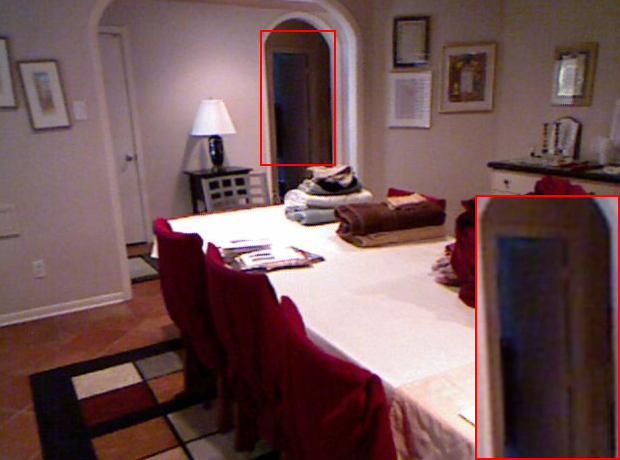}\vspace{2pt}
		\end{minipage}
	}	
	\hspace{-9pt}
	\subfloat[GT]{
		\begin{minipage}[c]{0.12\textwidth} 
			\includegraphics[width=1\textwidth]{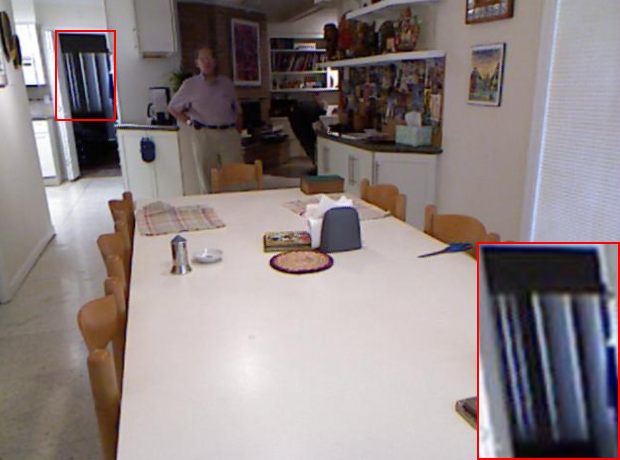}\vspace{2pt}
			\includegraphics[width=1\textwidth]{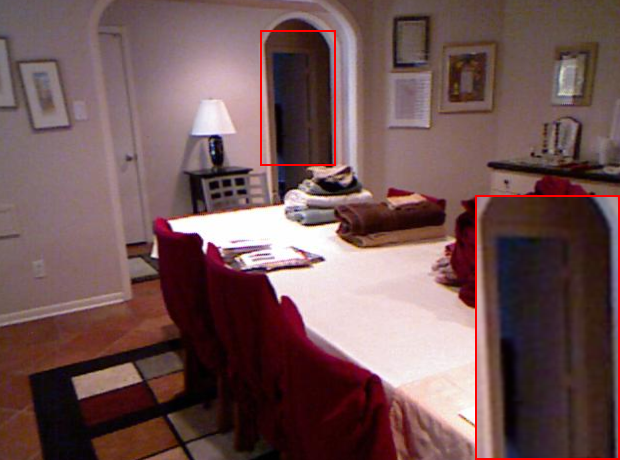}\vspace{2pt}
		\end{minipage}
	}	
	\caption{Visual comparisons of various methods on synthetic SOTS-indoor \cite{li2018TIP} dataset. Please zoom in on screen for a better view.}
	\label{fig:Comparison_ITS}
\end{figure*}

\begin{figure*}[p]
	\centering
	\subfloat[Hazy]{
		\begin{minipage}[c]{0.12\textwidth} 
			\includegraphics[width=1\textwidth]{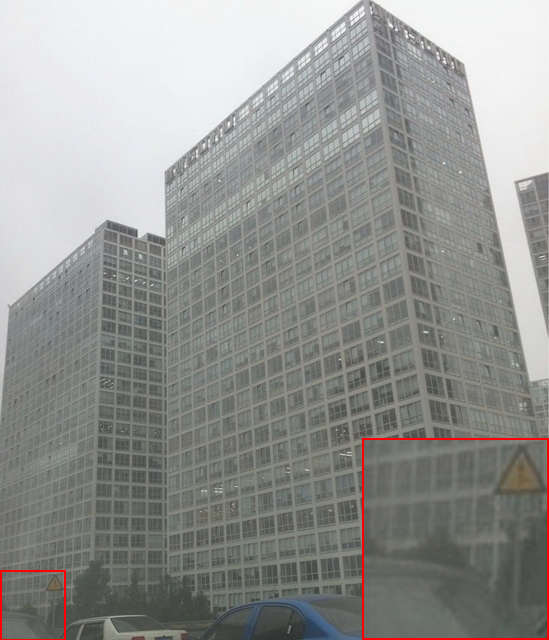}\vspace{2pt}
			\includegraphics[width=1\textwidth]{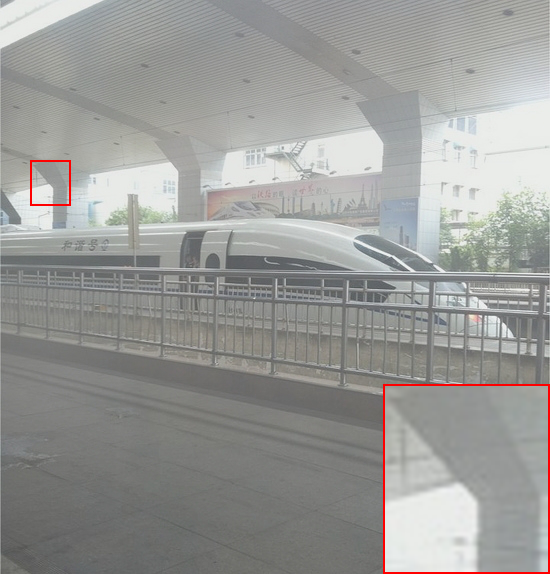}\vspace{2pt}
		\end{minipage}
	}	
	\hspace{-9pt}
	\subfloat[DCP \cite{he2010TPAMI}]{
		\begin{minipage}[c]{0.12\textwidth} 
			\includegraphics[width=1\textwidth]{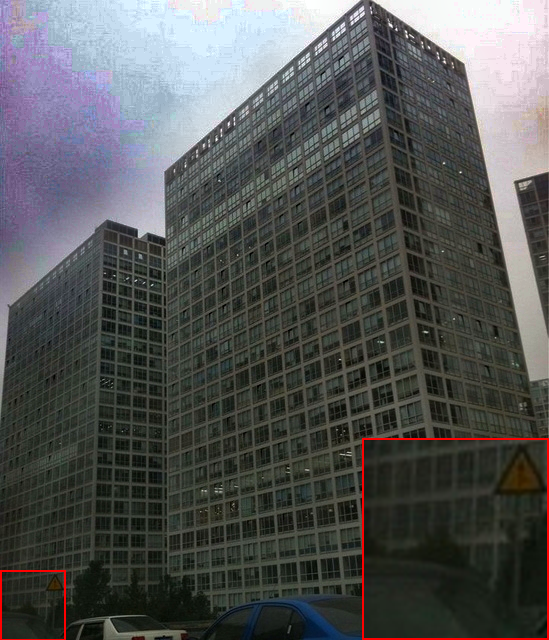}\vspace{2pt}
			\includegraphics[width=1\textwidth]{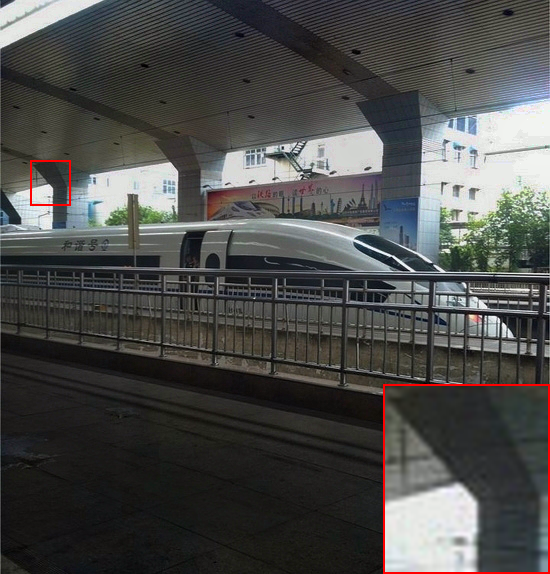}\vspace{2pt}
		\end{minipage}
	}	
	\hspace{-9pt}
	\subfloat[GDN \cite{liu2019ICCV}]{
		\begin{minipage}[c]{0.12\textwidth} 
			\includegraphics[width=1\textwidth]{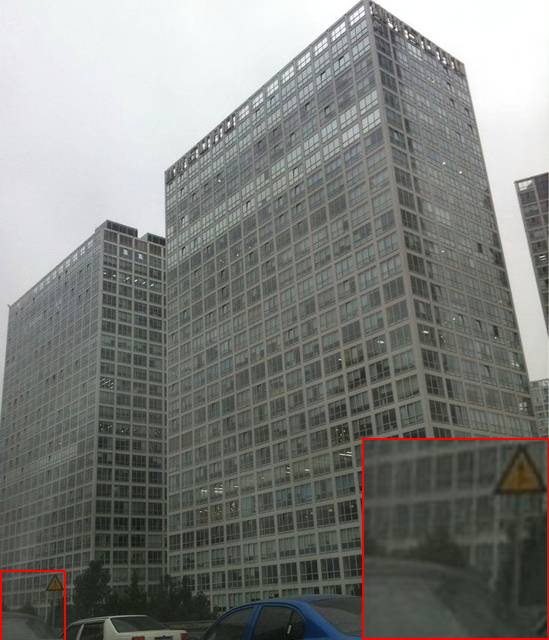}\vspace{2pt}
			\includegraphics[width=1\textwidth]{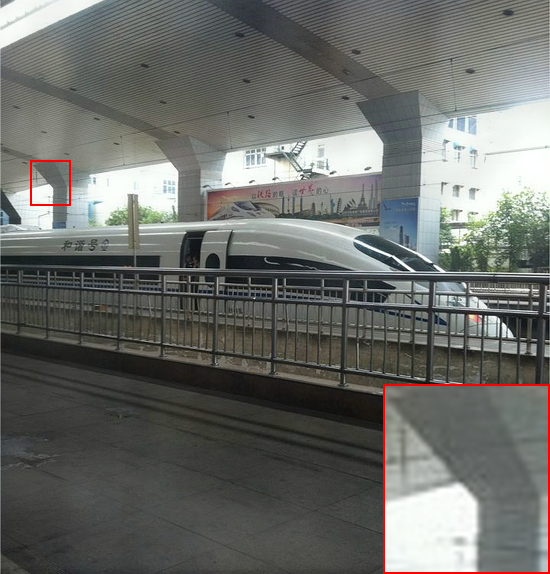}\vspace{2pt}
		\end{minipage}
	}	
	\hspace{-9pt}
	\subfloat[FFA \cite{qin2020AAAI}]{
		\begin{minipage}[c]{0.12\textwidth} 
			\includegraphics[width=1\textwidth]{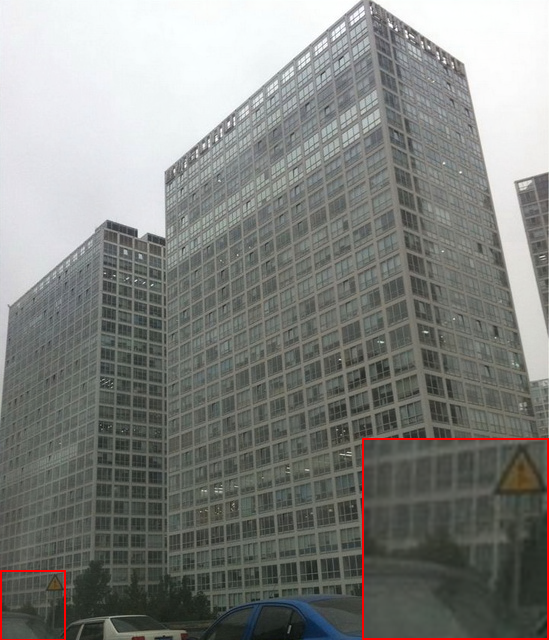}\vspace{2pt}
			\includegraphics[width=1\textwidth]{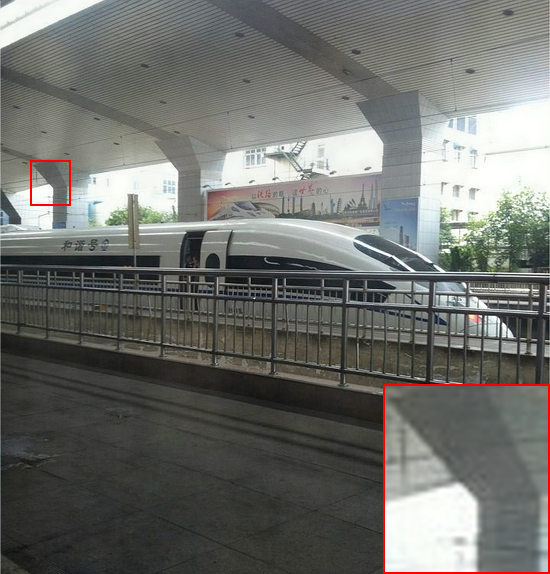}\vspace{2pt}
		\end{minipage}
	}	
	\hspace{-9pt}
	\subfloat[Dehamer \cite{guo2022CVPR}]{
		\begin{minipage}[c]{0.12\textwidth} 
			\includegraphics[width=1\textwidth]{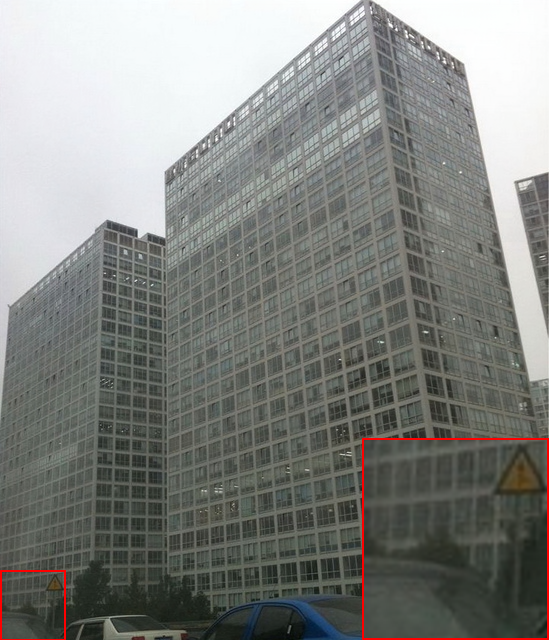}\vspace{2pt}
			\includegraphics[width=1\textwidth]{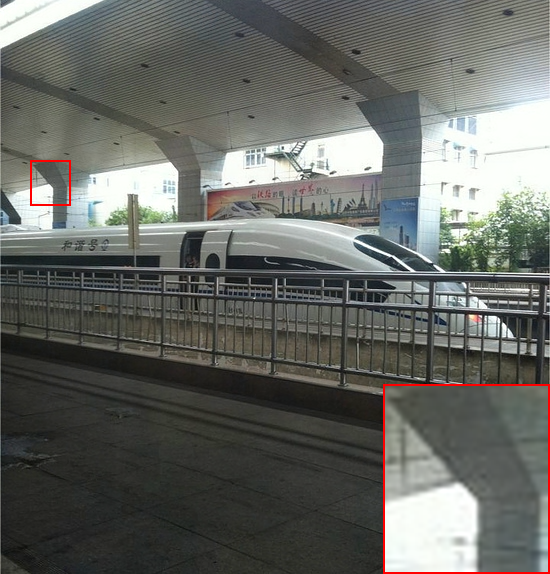}\vspace{2pt}
		\end{minipage}
	}	
	\hspace{-9pt}
	\subfloat[Ours]{
		\begin{minipage}[c]{0.12\textwidth} 
			\includegraphics[width=1\textwidth]{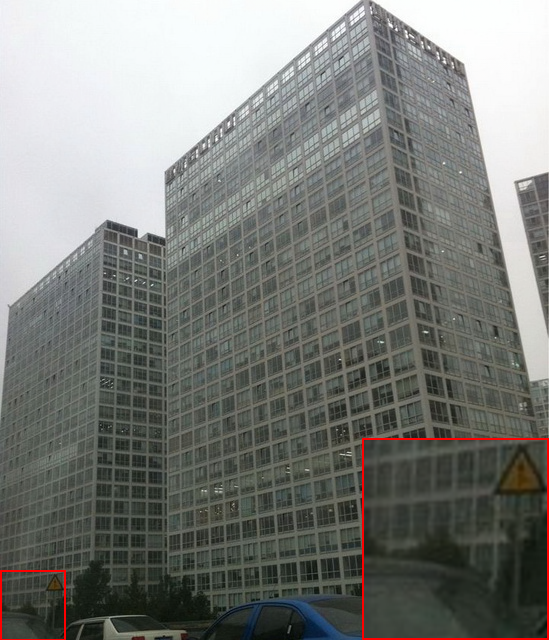}\vspace{2pt}
			\includegraphics[width=1\textwidth]{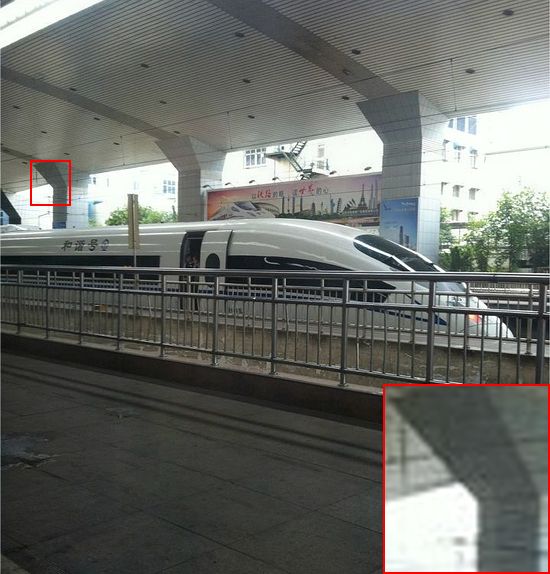}\vspace{2pt}
		\end{minipage}
	}	
	\hspace{-9pt}
	\subfloat[GT]{
		\begin{minipage}[c]{0.12\textwidth} 
			\includegraphics[width=1\textwidth]{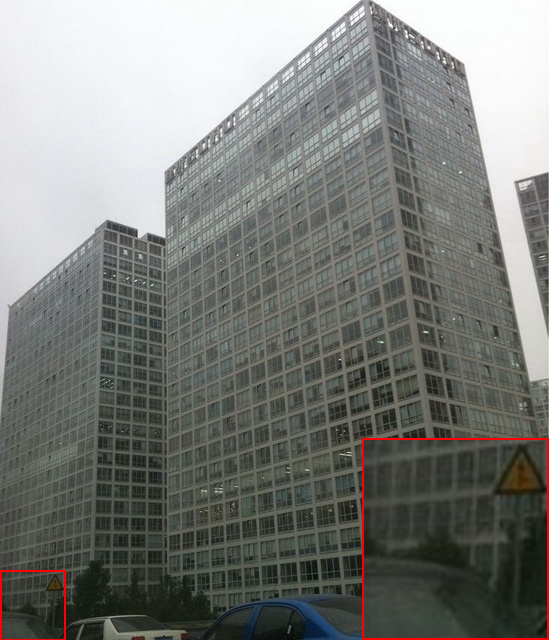}\vspace{2pt}
			\includegraphics[width=1\textwidth]{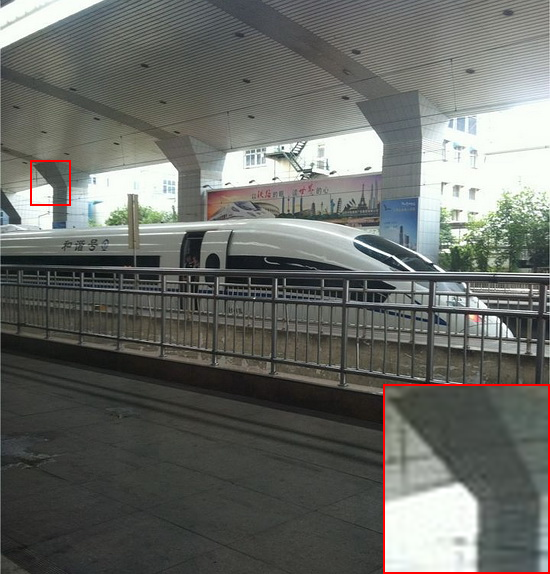}\vspace{2pt}
		\end{minipage}
	}	
	\caption{Visual comparisons of various methods on synthetic SOTS-outdoor \cite{li2018TIP} dataset. Please zoom in on screen for a better view.}
	\label{fig:Comparison_OTS}
\end{figure*}

\begin{figure*}[p]
	\centering
	\subfloat[Hazy]{
		\begin{minipage}[c]{0.12\textwidth} 
			\includegraphics[width=1\textwidth]{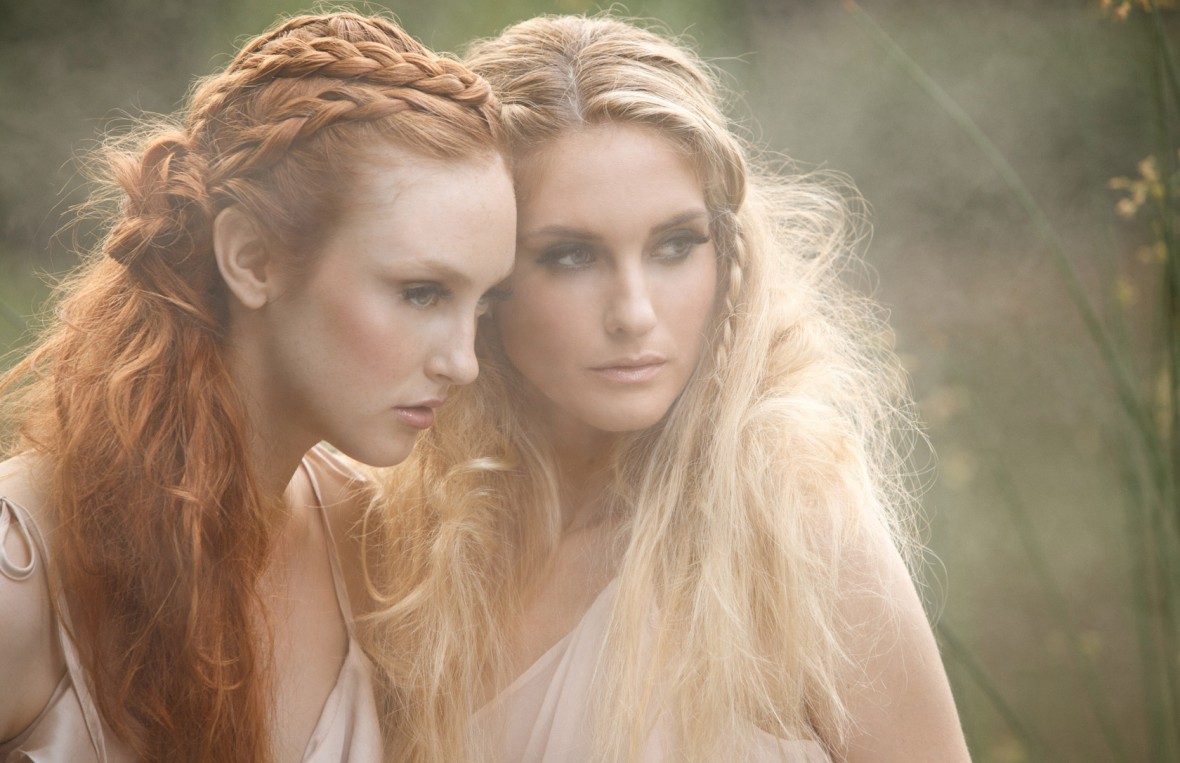}\vspace{2pt}
			\includegraphics[width=1\textwidth]{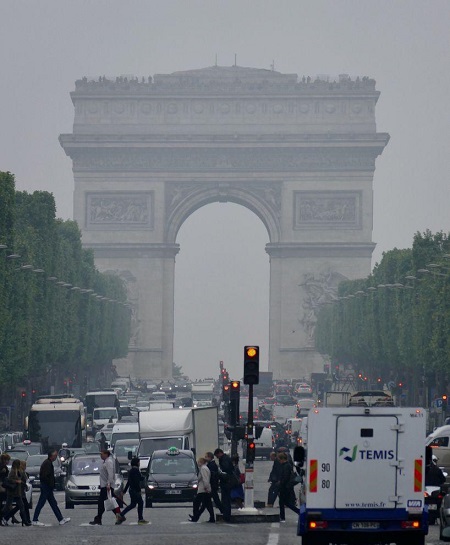}\vspace{2pt}
			\includegraphics[width=1\textwidth]{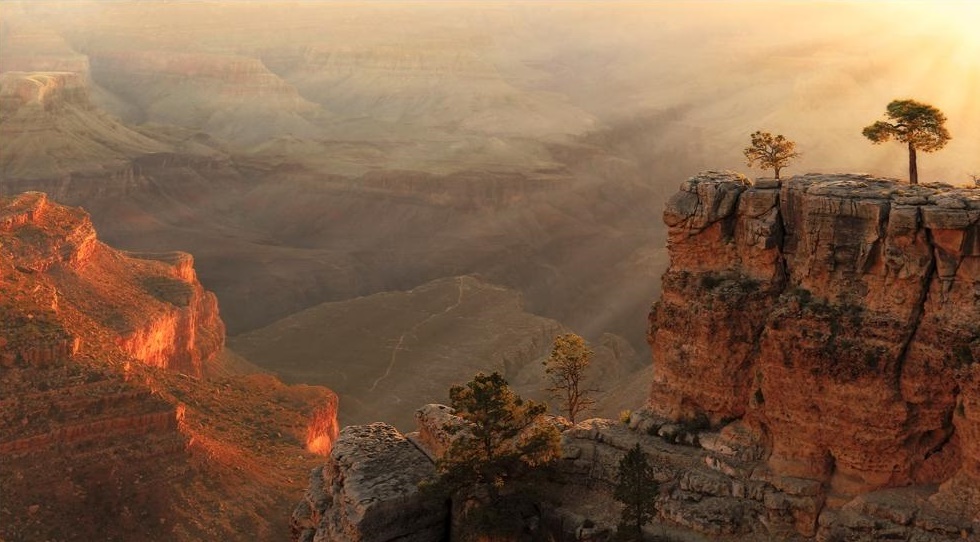}\vspace{2pt}
		\end{minipage}
	}	
	\hspace{-9pt}
	\subfloat[DCP \cite{he2010TPAMI}]{
		\begin{minipage}[c]{0.12\textwidth} 
			\includegraphics[width=1\textwidth]{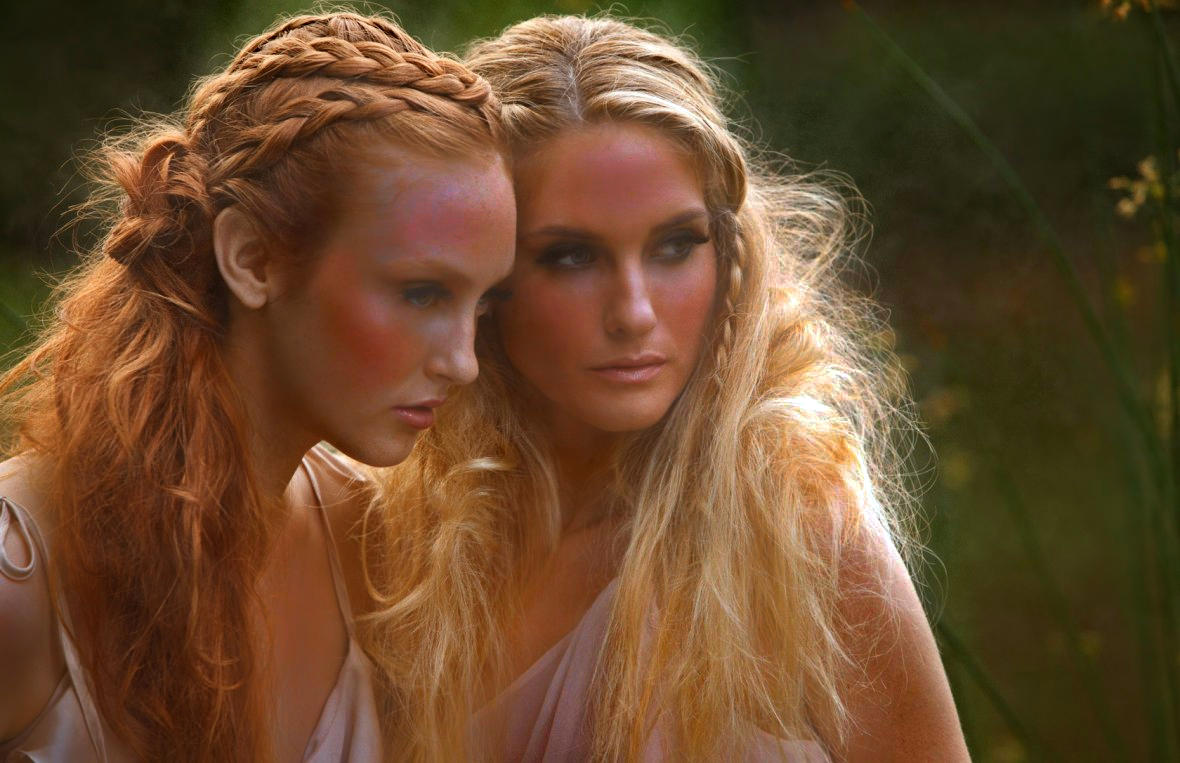}\vspace{2pt}
			\includegraphics[width=1\textwidth]{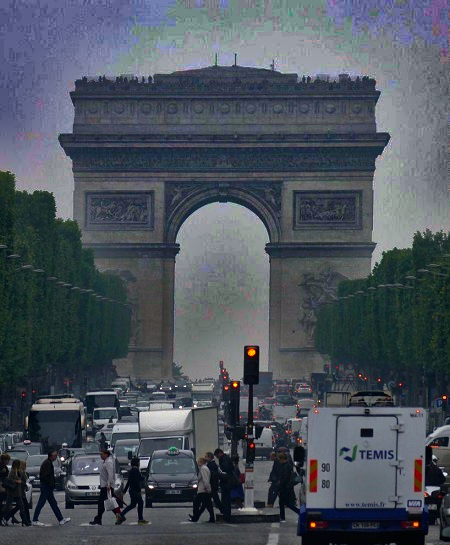}\vspace{2pt}
			\includegraphics[width=1\textwidth]{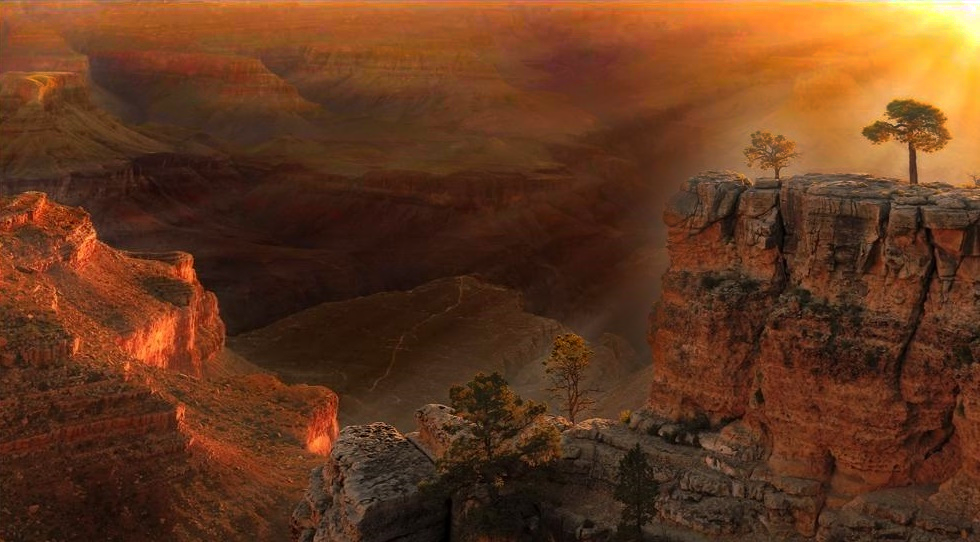}\vspace{2pt}
		\end{minipage}
	}	
	\hspace{-9pt}
	\subfloat[GDN \cite{liu2019ICCV}]{
		\begin{minipage}[c]{0.12\textwidth} 
			\includegraphics[width=1\textwidth]{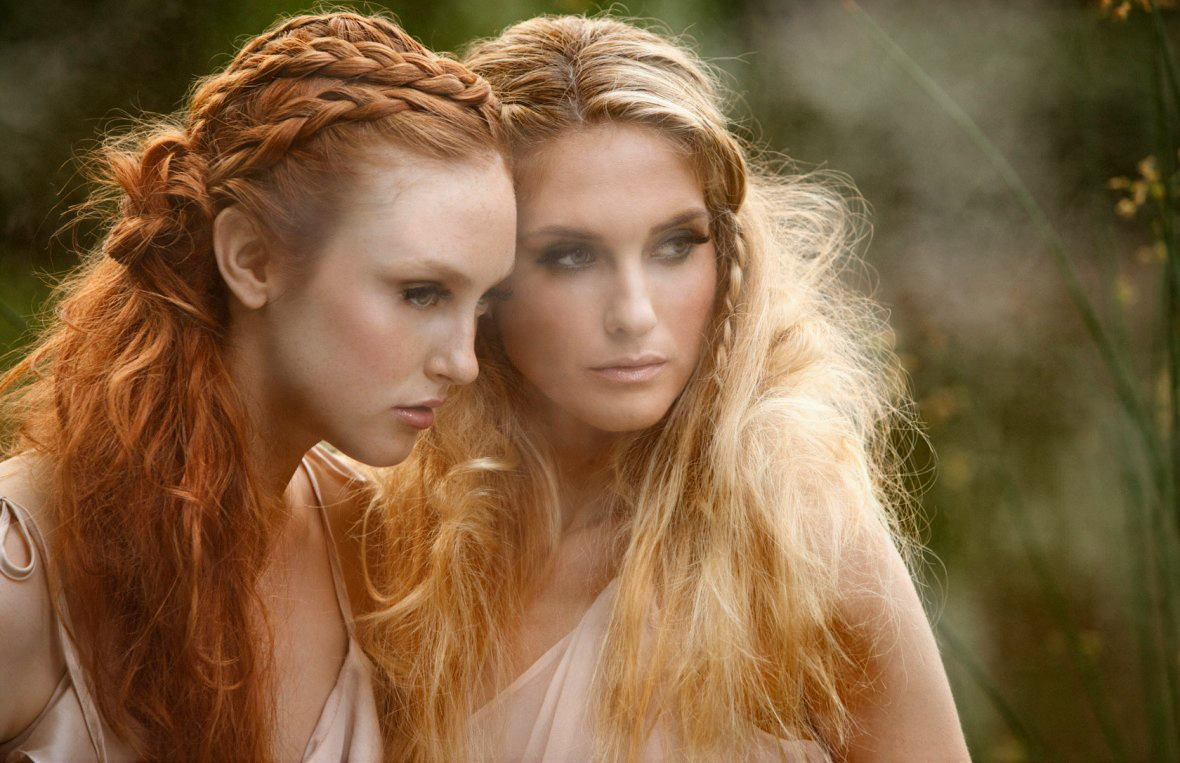}\vspace{2pt}
			\includegraphics[width=1\textwidth]{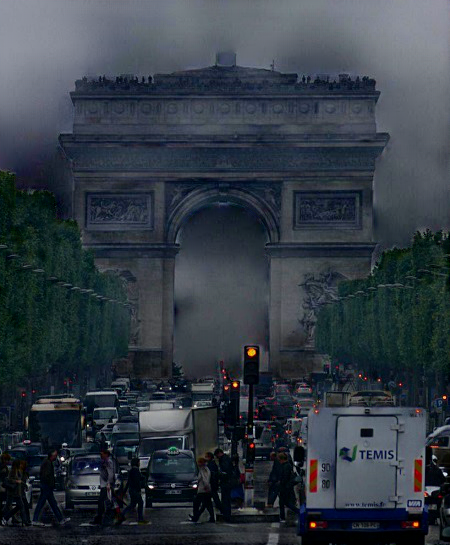}\vspace{2pt}
			\includegraphics[width=1\textwidth]{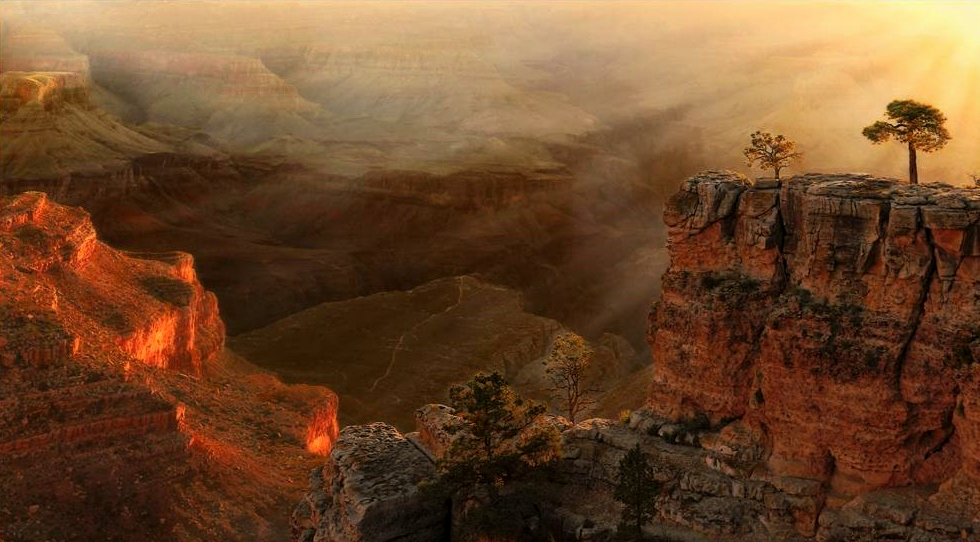}\vspace{2pt}
		\end{minipage}
	}	
	\hspace{-9pt}
	\subfloat[FFA \cite{qin2020AAAI}]{
		\begin{minipage}[c]{0.12\textwidth} 
			\includegraphics[width=1\textwidth]{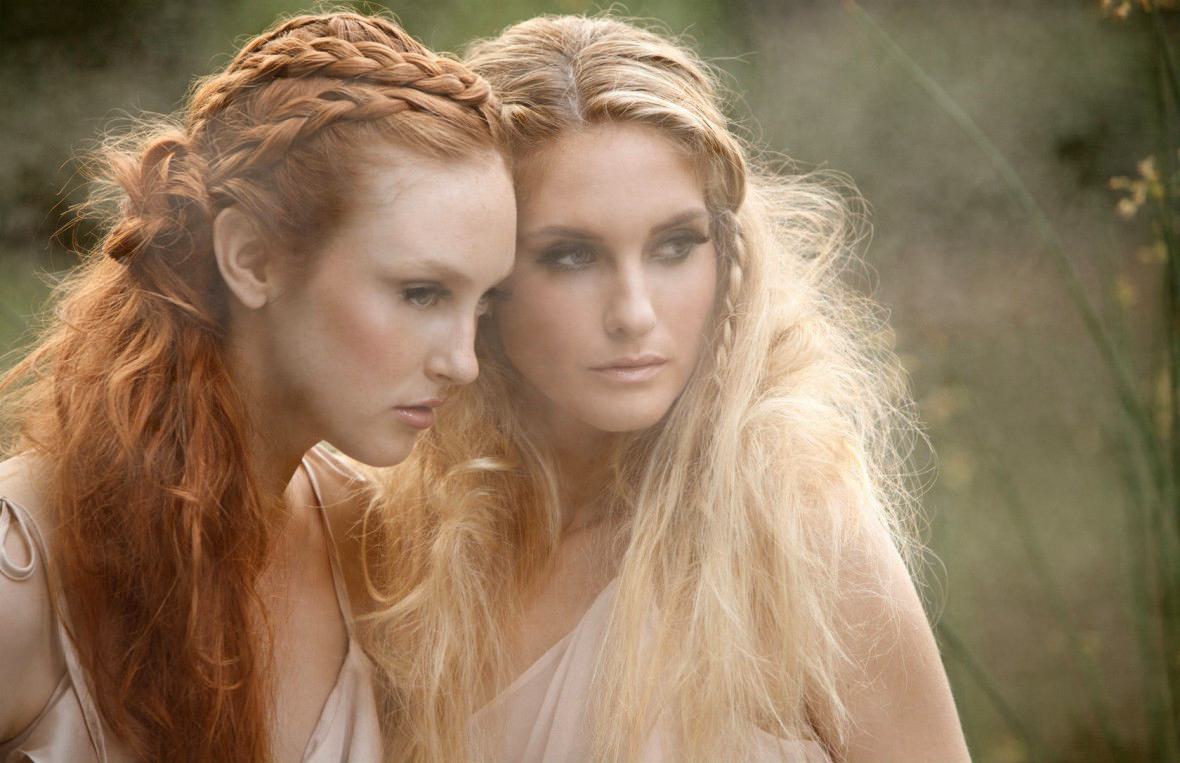}\vspace{2pt}
			\includegraphics[width=1\textwidth]{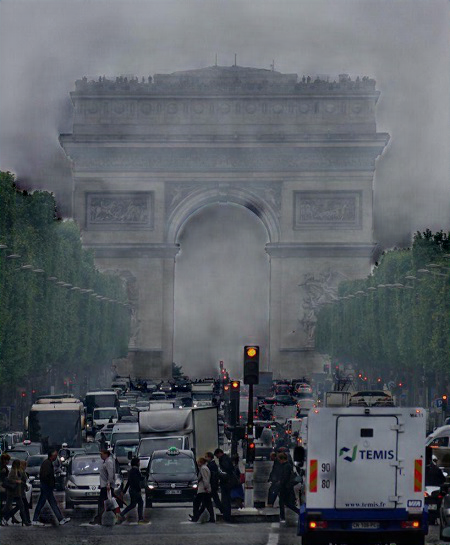}\vspace{2pt}
			\includegraphics[width=1\textwidth]{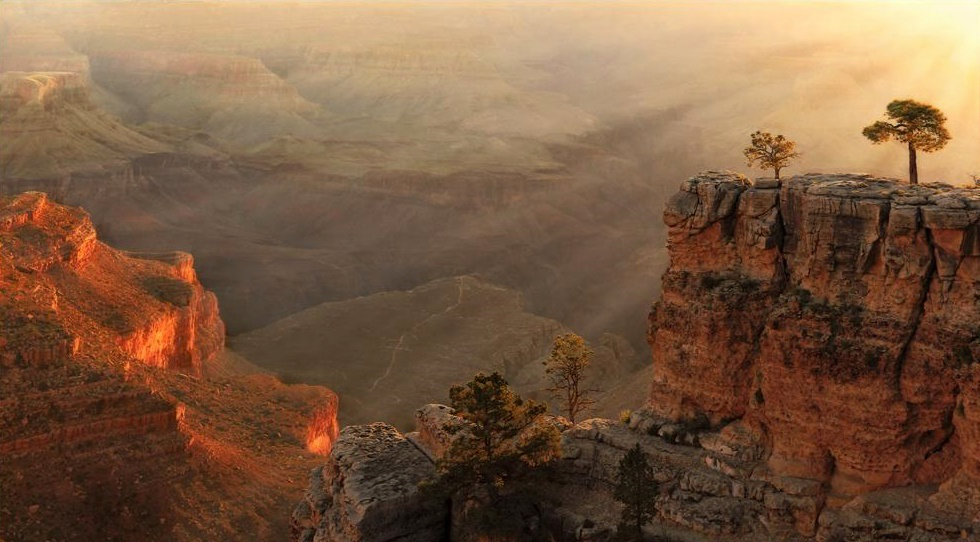}\vspace{2pt}
		\end{minipage}
	}	
	\hspace{-9pt}
	\subfloat[AECR-Net \cite{wu2021CVPR}]{
		\begin{minipage}[c]{0.12\textwidth} 
			\includegraphics[width=1\textwidth]{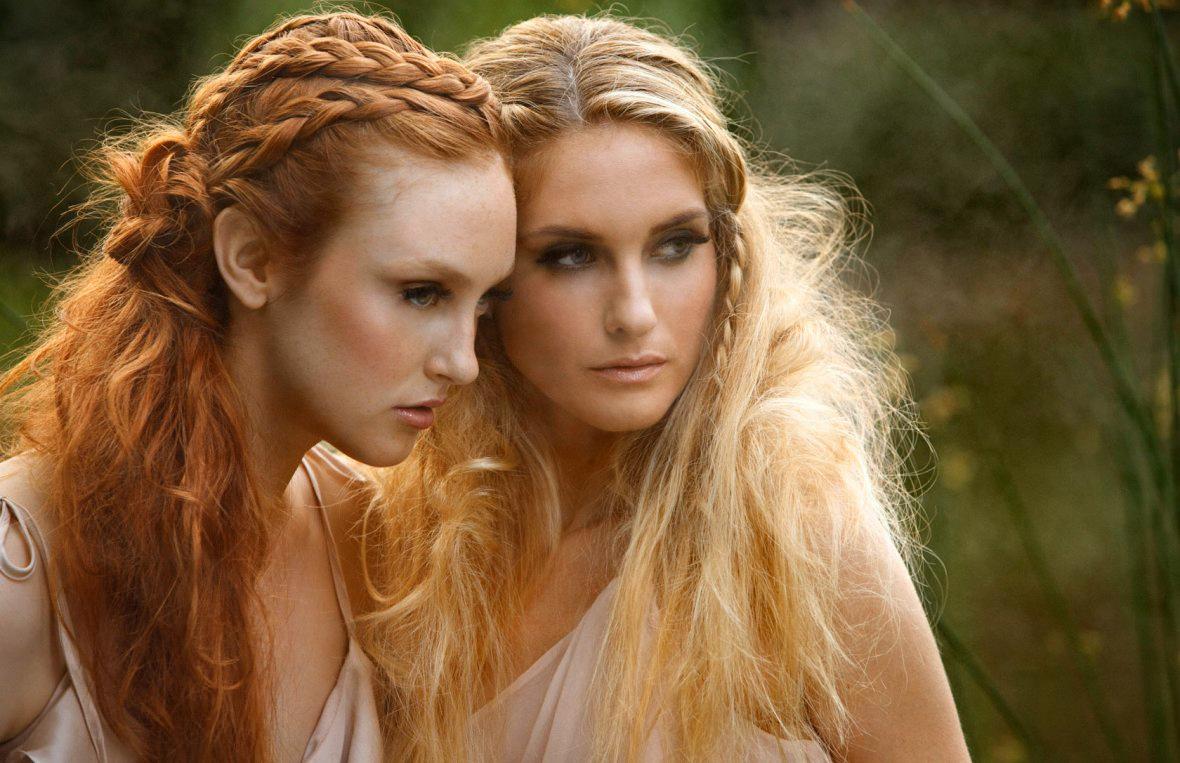}\vspace{2pt}
			\includegraphics[width=1\textwidth]{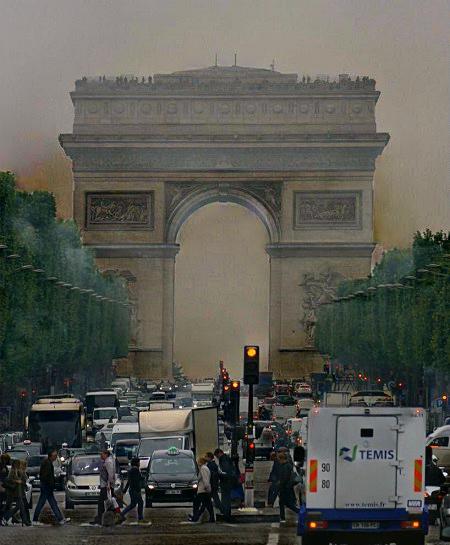}\vspace{2pt}
			\includegraphics[width=1\textwidth]{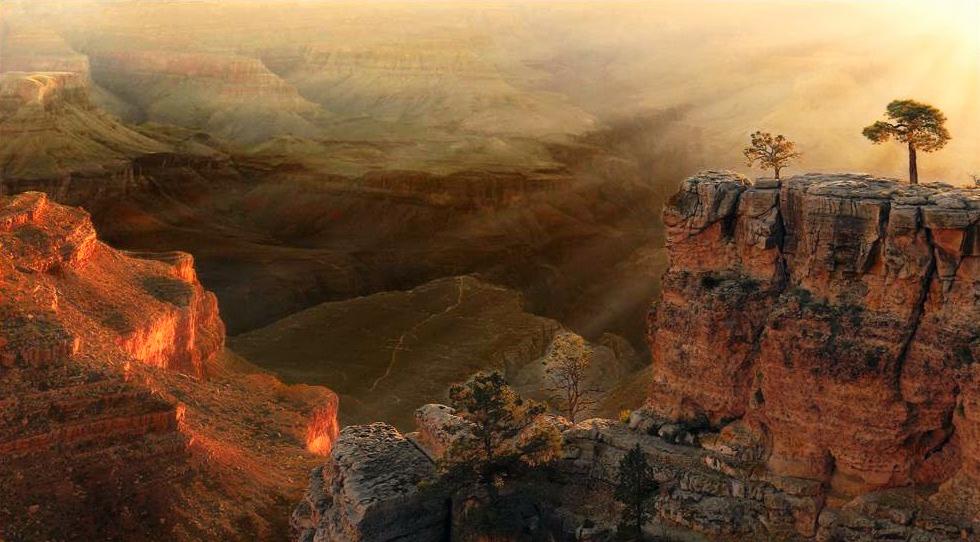}\vspace{2pt}
		\end{minipage}
	}	
	\hspace{-9pt}
	\subfloat[Dehamer \cite{guo2022CVPR}]{
		\begin{minipage}[c]{0.12\textwidth} 
			\includegraphics[width=1\textwidth]{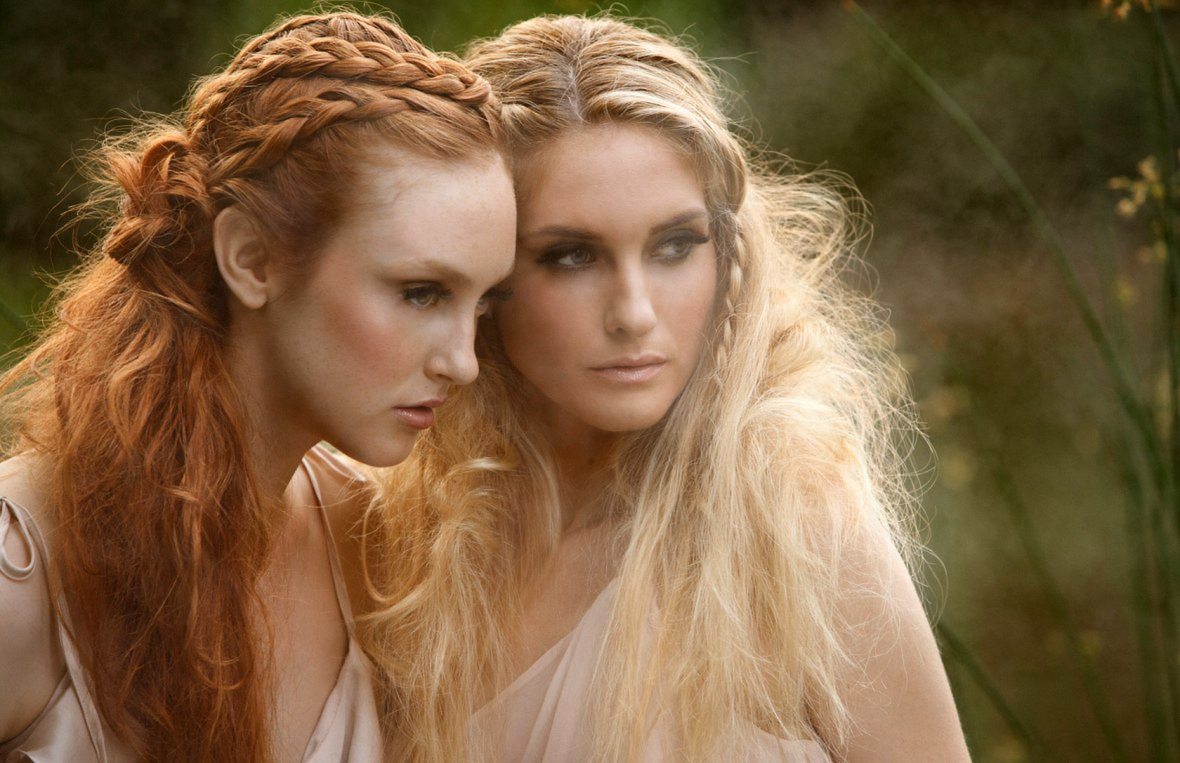}\vspace{2pt}
			\includegraphics[width=1\textwidth]{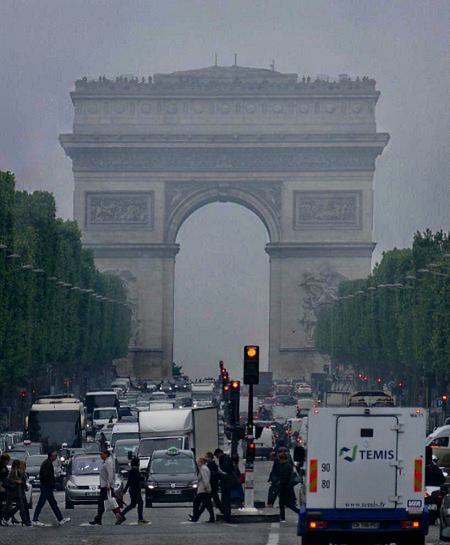}\vspace{2pt}
			\includegraphics[width=1\textwidth]{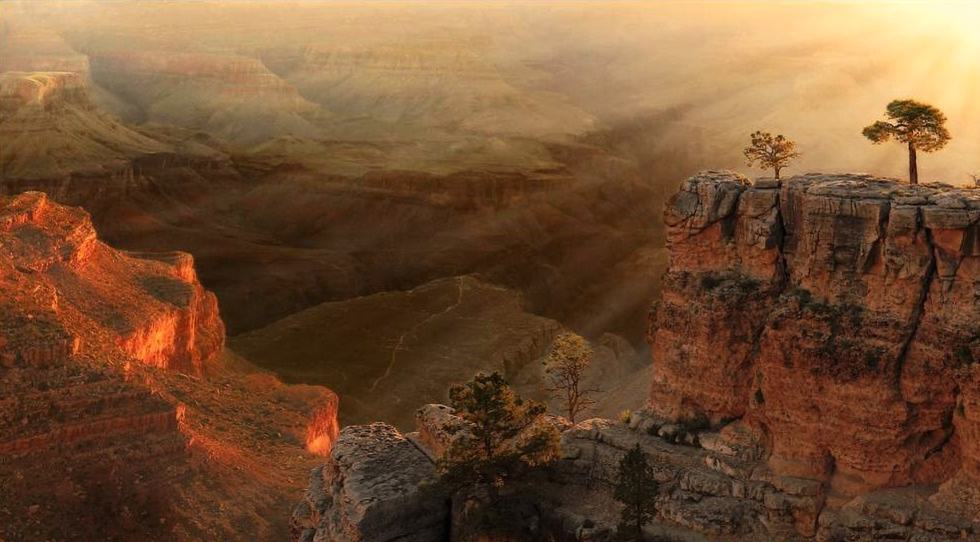}\vspace{2pt}
		\end{minipage}
	}	
	\hspace{-9pt}
	\subfloat[Ours]{
		\begin{minipage}[c]{0.12\textwidth} 
			\includegraphics[width=1\textwidth]{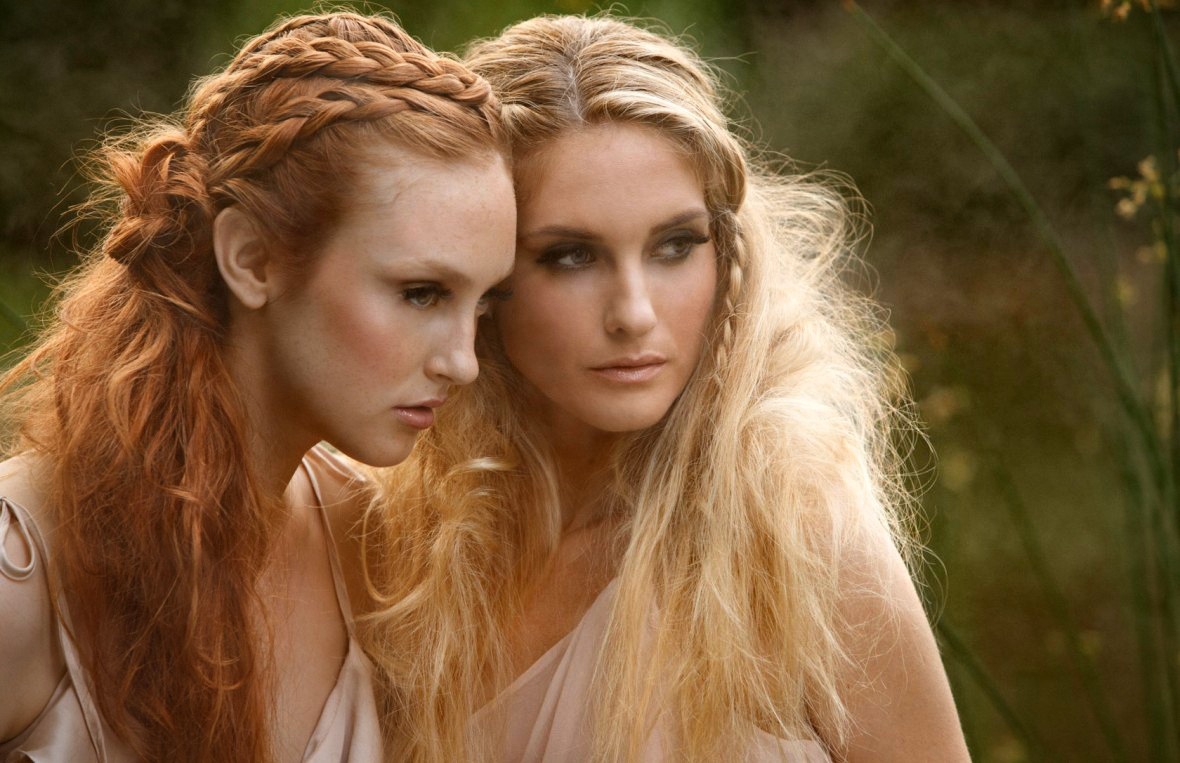}\vspace{2pt}
			\includegraphics[width=1\textwidth]{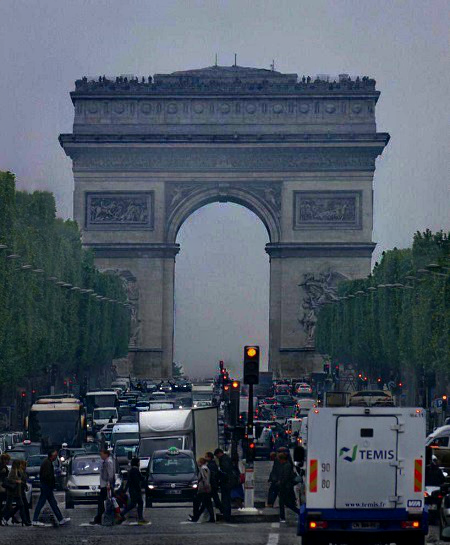}\vspace{2pt}
			\includegraphics[width=1\textwidth]{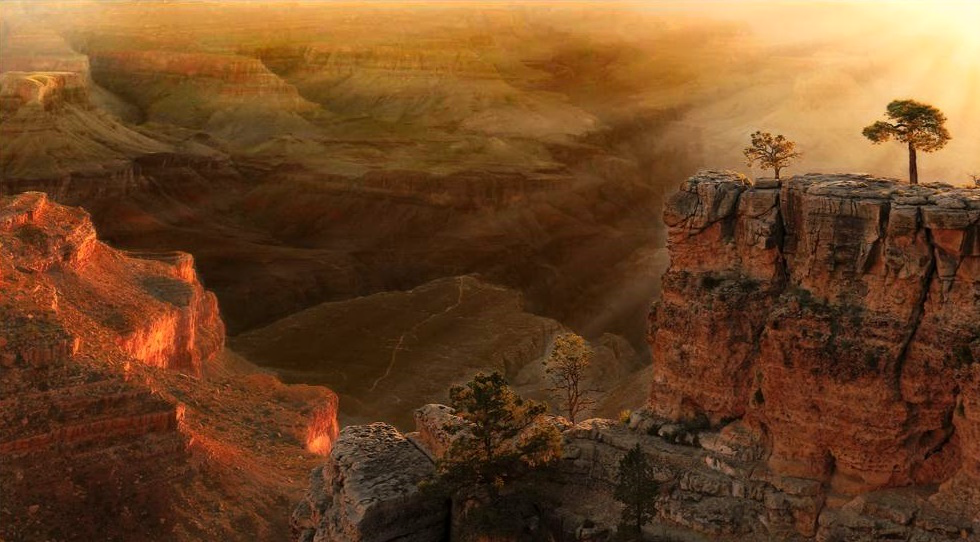}\vspace{2pt}
		\end{minipage}
	}	
	\caption{Dehazing results of various methods on real-world hazy images. Please zoom in on screen for a better view.}
	\label{fig:Comparison_Real}
\end{figure*}

\subsubsection[3]{CGA-based Mixup Fusion Scheme}
\label{subsec: CGA-based Mixup Fusion Scheme}
We further perform ablation study to verify the effectiveness of proposed CGA-based mixup fusion scheme.
We utilize \textit{Model\_DEAB} with mixup fusion scheme from AECR-Net \cite{wu2021CVPR} as the baseline, and then evaluate another two schemes: element-wise addition \cite{dong2020CVPR,bai2022TIP,ye2022ECCVORAL} and proposed CGA-based mixup.
Their models are referred to \textit{Model\_DEAB\_A} and \textit{Model\_DEAB\_C}.
The comparative results are shown in Table.~\ref{tab:table4}.
From these results, we see that addition achieves very similar performance with mixup (0.06 dB higher PSNR and 0.002 lower SSIM).
Addition is a special case of mixup with the constant weights, and we experimentally find that the initial values have a significant impact on the performance of mixup.
Note that, our proposed CGA-based mixup fusion scheme achieves the best performance in terms of PSNR and SSIM.

In addition, we deploy feature extraction blocks in level 1 and 2 to further improve the performance.
By deploying residual block (RB) in level 1 and level 2 (we refer this model to \textit{Model\_MS}), the performance improves by a large margin (2.52 dB in terms of PSNR).
It means transforming features in high-resolution space even full-resolution space can repair the lost information, which is critical for image regression.
Our final DEA-Net-S achieves 39.16 dB in terms of PSNR and 0.9921 in terms of SSIM by deploying DEB in level 1 and 2.
The suffix `-S' denotes the model is trained with the settings in ablation study, which is a simplified version.
For \textit{Model\_MS} and DEA-Net-S, $[N_1, N_2, N_3, N_4, N_5]$ is set to $[3, 3, 6, 3, 3]$.
It is worth mentioning that we omit the CGA in level 1 and level 2 (simplify DEAB into DEB) by taking the model complexity into account and to avoid complex hyper-parameter tuning (e.g., the reduction ratio).

\subsection{Comparisons with SOTA Methods}

In this section, we compare our proposed DEA-Net with 4 earlier dehazing approaches including DCP \cite{he2010TPAMI}, DehazeNet \cite{cai2016TIP}, AOD-Net \cite{li2017ICCV}, GFN \cite{ren2018CVPR} and 8 recent state-of-the-art (SOTA) single image dehazing methods including FFA-Net \cite{qin2020AAAI}, MSBDN \cite{dong2020CVPR}, DMT-Net \cite{liu2021ACMMM}, AECR-Net \cite{wu2021CVPR}, SGID-PFF \cite{bai2022TIP}, UDN \cite{hong2022AAAI}, PMDNet \cite{ye2022ECCVORAL}, Dehamer \cite{guo2022CVPR} on SOTS-Indoor, SOTS-Ourdoor, Haze4K datasets.
We report three DEA-Net variants including DEA-Net-S with the settings in ablation study (i.e., the final model of Table.~\ref{tab:table4}), DEA-Net with normal settings, and DEA-Net-CR with normal settings and the contrastive regularization (CR) from AECR-Net \cite{wu2021CVPR}.
DEA-Net-CR has identical setting of CR with AECR-Net \cite{wu2021CVPR}.
Note that CR will not increase additional parameters and inference time, since it can be directly removed in the testing phase.
For others, we adopt the official released codes or evaluation results of these methods for fair comparisons if they are publicly available, otherwise we retrain them using the same training datasets.

\textbf{Quantitative Analysis.}
Table.~\ref{tab:benchmark} shows quantitative evaluation results (PSNR and SSIM indexes) of our DEA-Nets and other state-of-the-art methods on SOTS \cite{li2018TIP} and Haze4K \cite{liu2021ACMMM}.
As we can see, even our DEA-Net-S achieves the best performance with 39.16 dB PSNR and 0.9921 SSIM on SOTS-indoor than the alternatives.
Further, our DEA-Net and DEA-Net-CR improve the performance by a large margin on both SOTS-indoor and SOTS-outdoor.
On Haze4K dataset, our DEA-Net and DEA-Net-CR achieve the best SSIM (0.9869 and 0.9885). 
We round the results to two decimals to keep consistent with \cite{liu2021ACMMM}.
Our DEA-Net-CR ranks first in all comparisons on SOTS and Haze4K.

In addition, we adopt number of parameters (\# Param.), number of floating-point operations (\# FLOPs), and runtime as the major indicators of computational efficiency.
The earlier dehazing methods contain very small parameters sizes at the cost of a big performance drop.
Compared with recent SOTA methods, our DEA-Nets run fastest with acceptable \# Param. and \# FLOPs.
Any one of our DEA-Net variants can rank second best in terms of \# Param. and \# FLOPs.
This implies our DEA-Nets can reach a good trade-off between performance and model complexity.
Note that \# FLOPs and runtime are measured on color images with $256\times 256$ resolution.


\textbf{Qualitative Analysis.}
Fig.~\ref{fig:Comparison_ITS} shows the visual comparisons between our DEA-Net and previous SOTA methods on synthetic SOTS-indoor dataset.
Our proposed DEA-Net can recover sharper and clearer contours or edges, and the results obtained by DEA-Net contains less haze residuals.
Fig.~\ref{fig:Comparison_OTS} shows the visual comparisons on synthetic SOTS-ourdoor dataset.
We observe that in outdoor scenes, the results of our DEA-Net are closest to the ground truth than the other alternatives.
We also test our DEA-Net on real-world hazy images, and compare the results with various SOTA methods.
As shown, the other methods either remain haze on the processed results or produce color deviations and artifacts.
On the contrary, our DEA-Net can output more visually pleasing dehazing results.

\section{Conclusion}
\label{sec: conclusion}
In this paper, we propose a DEA-Net to deal with the challenging single image dehazing problem.
Specifically, we design the detail-enhanced convolution (DEConv) by introducing the difference convolution to integrate local descriptors into normal convolution layer.
Compare with vanilla convolution, DEConv has enhanced representation and generalization capacity.
In addition, the DEConv can be equivalently converted into a vanilla convolution without triggering extra parameters and computational cost.
Then, we design a sophisticated attention mechanism termed content-guided attention (CGA), which assigns unique spatial importance map (SIM) to every channel.
With CGA, more useful information encoded in features can be emphasized.
Based on CGA, we further present a fusion scheme to effectively fuse low-level features in the encoder part with corresponding high-level features.
Extensive experiments show that our DEA-Net achieves state-of-the-art results quantitatively and qualitatively.



\bibliographystyle{IEEEtran}
\bibliography{citations}

\end{document}